\def\cleancopy{}
\documentclass[journal]{IEEEtran}

\usepackage[utf8]{inputenc}
\usepackage[T1]{fontenc}
\usepackage{graphicx}
\graphicspath{{figures/}}
\usepackage{booktabs}
\usepackage{multirow}
\usepackage{amsmath,amssymb}
\usepackage[table]{xcolor}
\usepackage{subcaption}
\usepackage{cite}
\usepackage{url}
\usepackage[hidelinks]{hyperref}
\usepackage{siunitx}
\usepackage{float,longtable}
\newcommand{\rsq}{$R^2$}
\ifdefined\cleancopy\newcommand{\todo}[1]{}\else\newcommand{\todo}[1]{\textcolor{red}{[#1]}}\fi
\newcommand{\hglicf}{0.727}
\newcommand{\hglicfsd}{0.014}
\newcommand{\hglicfml}{0.719}
\newcommand{\hglicfap}{0.667}

\newcommand{\hglicfg}{0.763}
\newcommand{\hglicfgsd}{0.012}
\newcommand{\hglicfgml}{0.770}
\newcommand{\hglicfgap}{0.729}

\newcommand{\hglcnnlstm}{0.804}
\newcommand{\hglcnnlstmsd}{0.003}
\newcommand{\hglcnnlstmml}{0.775}
\newcommand{\hglcnnlstmap}{0.769}

\newcommand{\hglcnn}{0.760}
\newcommand{\hglcnnsd}{0.010}
\newcommand{\hglcnnml}{0.784}
\newcommand{\hglcnnap}{0.769}

\newcommand{\pglcnnlstmicf}{0.031}
\newcommand{\pglcnnlstmicfd}{+0.065}

\newcommand{\hphicf}{0.648}
\newcommand{\hphicfsd}{0.015}
\newcommand{\hphicfml}{0.451}
\newcommand{\hphicfap}{0.485}

\newcommand{\hphicfg}{0.642}
\newcommand{\hphicfgsd}{0.003}
\newcommand{\hphicfgml}{0.435}
\newcommand{\hphicfgap}{0.483}

\newcommand{\hphcnnlstm}{0.622}
\newcommand{\hphcnnlstmsd}{0.007}
\newcommand{\hphcnnlstmml}{0.391}
\newcommand{\hphcnnlstmap}{0.458}

\newcommand{\hphcnn}{0.583}
\newcommand{\hphcnnsd}{0.008}
\newcommand{\hphcnnml}{0.358}
\newcommand{\hphcnnap}{0.441}

\newcommand{\pphcnnlstmicf}{0.131}
\newcommand{\pphcnnlstmicfd}{-0.031}

\newcommand{\seedpending}{}
\newcommand{\tsglicf}{0.791}
\newcommand{\inflglicf}{+0.064}
\newcommand{\tsglcnnlstm}{0.894}
\newcommand{\inflglcnnlstm}{+0.090}
\newcommand{\tsglcnn}{0.800}
\newcommand{\inflglcnn}{+0.040}
\newcommand{\tsphicf}{0.790}
\newcommand{\inflphicf}{+0.142}
\newcommand{\tsphcnnlstm}{0.790}
\newcommand{\inflphcnnlstm}{+0.168}
\newcommand{\tsphcnn}{0.777}
\newcommand{\inflphcnn}{+0.194}
\newcommand{\trialpending}{}
\newcommand{\seedsd}{SD across three seeds of the fold mean}
\newcommand{\ablglbaseliner}{0.590}
\newcommand{\ablglbaselineml}{0.682}
\newcommand{\ablglbaselinevert}{0.816}
\newcommand{\ablglbaselineap}{0.108}
\newcommand{\ablglbaselinepx}{0.808}
\newcommand{\ablglbaselinepz}{0.536}
\newcommand{\ablglcorotr}{0.713}
\newcommand{\ablglcorotml}{0.688}
\newcommand{\ablglcorotvert}{0.828}
\newcommand{\ablglcorotap}{0.675}
\newcommand{\ablglcorotpx}{0.808}
\newcommand{\ablglcorotpz}{0.566}
\newcommand{\ablglperfoldr}{0.608}
\newcommand{\ablglperfoldml}{0.716}
\newcommand{\ablglperfoldvert}{0.862}
\newcommand{\ablglperfoldap}{0.057}
\newcommand{\ablglperfoldpx}{0.838}
\newcommand{\ablglperfoldpz}{0.570}
\newcommand{\ablglbothr}{0.732}
\newcommand{\ablglbothml}{0.722}
\newcommand{\ablglbothvert}{0.854}
\newcommand{\ablglbothap}{0.700}
\newcommand{\ablglbothpx}{0.834}
\newcommand{\ablglbothpz}{0.550}
\newcommand{\ablphbaseliner}{0.588}
\newcommand{\ablphbaselineml}{0.483}
\newcommand{\ablphbaselinevert}{0.664}
\newcommand{\ablphbaselineap}{0.397}
\newcommand{\ablphbaselinepx}{0.791}
\newcommand{\ablphbaselinepz}{0.607}
\newcommand{\ablphcorotr}{0.583}
\newcommand{\ablphcorotml}{0.420}
\newcommand{\ablphcorotvert}{0.657}
\newcommand{\ablphcorotap}{0.448}
\newcommand{\ablphcorotpx}{0.796}
\newcommand{\ablphcorotpz}{0.594}
\newcommand{\ablphperfoldr}{0.636}
\newcommand{\ablphperfoldml}{0.507}
\newcommand{\ablphperfoldvert}{0.700}
\newcommand{\ablphperfoldap}{0.386}
\newcommand{\ablphperfoldpx}{0.880}
\newcommand{\ablphperfoldpz}{0.705}
\newcommand{\ablphbothr}{0.627}
\newcommand{\ablphbothml}{0.417}
\newcommand{\ablphbothvert}{0.697}
\newcommand{\ablphbothap}{0.430}
\newcommand{\ablphbothpx}{0.892}
\newcommand{\ablphbothpz}{0.697}
\newcommand{\ablcnnlstmglbaseliner}{0.648}
\newcommand{\ablcnnlstmglbaselineml}{0.698}
\newcommand{\ablcnnlstmglbaselinevert}{0.856}
\newcommand{\ablcnnlstmglbaselineap}{0.228}
\newcommand{\ablcnnlstmglbaselinepx}{0.810}
\newcommand{\ablcnnlstmglbaselinepz}{0.650}
\newcommand{\ablcnnlstmglcorotr}{0.721}
\newcommand{\ablcnnlstmglcorotml}{0.695}
\newcommand{\ablcnnlstmglcorotvert}{0.835}
\newcommand{\ablcnnlstmglcorotap}{0.653}
\newcommand{\ablcnnlstmglcorotpx}{0.794}
\newcommand{\ablcnnlstmglcorotpz}{0.628}
\newcommand{\ablcnnlstmglperfoldr}{0.695}
\newcommand{\ablcnnlstmglperfoldml}{0.779}
\newcommand{\ablcnnlstmglperfoldvert}{0.905}
\newcommand{\ablcnnlstmglperfoldap}{0.246}
\newcommand{\ablcnnlstmglperfoldpx}{0.842}
\newcommand{\ablcnnlstmglperfoldpz}{0.705}
\newcommand{\ablcnnlstmglbothr}{0.801}
\newcommand{\ablcnnlstmglbothml}{0.791}
\newcommand{\ablcnnlstmglbothvert}{0.895}
\newcommand{\ablcnnlstmglbothap}{0.766}
\newcommand{\ablcnnlstmglbothpx}{0.832}
\newcommand{\ablcnnlstmglbothpz}{0.723}
\newcommand{\ablcnnlstmphbaseliner}{0.589}
\newcommand{\ablcnnlstmphbaselineml}{0.490}
\newcommand{\ablcnnlstmphbaselinevert}{0.662}
\newcommand{\ablcnnlstmphbaselineap}{0.395}
\newcommand{\ablcnnlstmphbaselinepx}{0.814}
\newcommand{\ablcnnlstmphbaselinepz}{0.586}
\newcommand{\ablcnnlstmphcorotr}{0.569}
\newcommand{\ablcnnlstmphcorotml}{0.393}
\newcommand{\ablcnnlstmphcorotvert}{0.647}
\newcommand{\ablcnnlstmphcorotap}{0.432}
\newcommand{\ablcnnlstmphcorotpx}{0.804}
\newcommand{\ablcnnlstmphcorotpz}{0.569}
\newcommand{\ablcnnlstmphperfoldr}{0.620}
\newcommand{\ablcnnlstmphperfoldml}{0.492}
\newcommand{\ablcnnlstmphperfoldvert}{0.730}
\newcommand{\ablcnnlstmphperfoldap}{0.371}
\newcommand{\ablcnnlstmphperfoldpx}{0.822}
\newcommand{\ablcnnlstmphperfoldpz}{0.686}
\newcommand{\ablcnnlstmphbothr}{0.614}
\newcommand{\ablcnnlstmphbothml}{0.376}
\newcommand{\ablcnnlstmphbothvert}{0.723}
\newcommand{\ablcnnlstmphbothap}{0.489}
\newcommand{\ablcnnlstmphbothpx}{0.801}
\newcommand{\ablcnnlstmphbothpz}{0.682}
\newcommand{\ablcnnglbaseliner}{0.603}
\newcommand{\ablcnnglbaselineml}{0.677}
\newcommand{\ablcnnglbaselinevert}{0.788}
\newcommand{\ablcnnglbaselineap}{0.268}
\newcommand{\ablcnnglbaselinepx}{0.747}
\newcommand{\ablcnnglbaselinepz}{0.536}
\newcommand{\ablcnnglcorotr}{0.683}
\newcommand{\ablcnnglcorotml}{0.671}
\newcommand{\ablcnnglcorotvert}{0.781}
\newcommand{\ablcnnglcorotap}{0.686}
\newcommand{\ablcnnglcorotpx}{0.748}
\newcommand{\ablcnnglcorotpz}{0.529}
\newcommand{\ablcnnglperfoldr}{0.657}
\newcommand{\ablcnnglperfoldml}{0.772}
\newcommand{\ablcnnglperfoldvert}{0.888}
\newcommand{\ablcnnglperfoldap}{0.254}
\newcommand{\ablcnnglperfoldpx}{0.824}
\newcommand{\ablcnnglperfoldpz}{0.546}
\newcommand{\ablcnnglbothr}{0.760}
\newcommand{\ablcnnglbothml}{0.784}
\newcommand{\ablcnnglbothvert}{0.884}
\newcommand{\ablcnnglbothap}{0.761}
\newcommand{\ablcnnglbothpx}{0.820}
\newcommand{\ablcnnglbothpz}{0.553}
\newcommand{\ablcnnphbaseliner}{0.516}
\newcommand{\ablcnnphbaselineml}{0.445}
\newcommand{\ablcnnphbaselinevert}{0.577}
\newcommand{\ablcnnphbaselineap}{0.332}
\newcommand{\ablcnnphbaselinepx}{0.694}
\newcommand{\ablcnnphbaselinepz}{0.534}
\newcommand{\ablcnnphcorotr}{0.523}
\newcommand{\ablcnnphcorotml}{0.356}
\newcommand{\ablcnnphcorotvert}{0.593}
\newcommand{\ablcnnphcorotap}{0.427}
\newcommand{\ablcnnphcorotpx}{0.693}
\newcommand{\ablcnnphcorotpz}{0.543}
\newcommand{\ablcnnphperfoldr}{0.606}
\newcommand{\ablcnnphperfoldml}{0.519}
\newcommand{\ablcnnphperfoldvert}{0.697}
\newcommand{\ablcnnphperfoldap}{0.387}
\newcommand{\ablcnnphperfoldpx}{0.768}
\newcommand{\ablcnnphperfoldpz}{0.658}
\newcommand{\ablcnnphbothr}{0.577}
\newcommand{\ablcnnphbothml}{0.355}
\newcommand{\ablcnnphbothvert}{0.684}
\newcommand{\ablcnnphbothap}{0.451}
\newcommand{\ablcnnphbothpx}{0.714}
\newcommand{\ablcnnphbothpz}{0.681}
\newcommand{\ablcnnlstmglapgain}{0.43}
\newcommand{\ablcnnlstmphmldrop}{-0.10}
\newcommand{\ablcnnglapgain}{0.42}

\newcommand{\ablphmldrop}{-0.06}
\newcommand{\ablpairedgldelta}{+0.142}
\newcommand{\ablpairedglwins}{6/6}
\newcommand{\ablpairedglp}{0.031}

\newcommand{\xlglphicf}{-3.73}
\newcommand{\xlglphicfvert}{-0.71}

\newcommand{\xlphglicf}{-1.08}
\newcommand{\xlphglicfvert}{-0.58}

\newcommand{\xlglphcnnlstm}{-1.91}
\newcommand{\xlglphcnnlstmvert}{-0.65}

\newcommand{\xlphglcnnlstm}{-0.35}
\newcommand{\xlphglcnnlstmvert}{0.12}

\newcommand{\xlglphcnn}{$<-10$}
\newcommand{\xlglphcnnvert}{$<-10$}

\newcommand{\xlphglcnn}{$<-10$}
\newcommand{\xlphglcnnvert}{$<-10$}

\newcommand{\xlsubglphicf}{-0.45}
\newcommand{\xlsubglphcnnlstm}{-0.78}
\newcommand{\xlsubglphcnn}{-0.72}
\newcommand{\xlsubphglicf}{-1.40}
\newcommand{\xlsubphglcnnlstm}{-1.49}
\newcommand{\xlsubphglcnn}{-1.10}
\newcommand{\combicfgl}{0.784}
\newcommand{\combicfglsd}{0.066}
\newcommand{\combicfph}{0.715}
\newcommand{\combicfphsd}{0.058}
\newcommand{\combcnnlstmgl}{0.841}
\newcommand{\combcnnlstmglsd}{0.021}
\newcommand{\combcnnlstmph}{0.718}
\newcommand{\combcnnlstmphsd}{0.070}
\newcommand{\combcnngl}{0.758}
\newcommand{\combcnnglsd}{0.033}
\newcommand{\combcnnph}{0.675}
\newcommand{\combcnnphsd}{0.080}

\newcommand{\maticfgldelta}{-0.008}
\newcommand{\maticfglwins}{3/6}

\newcommand{\maticfphdelta}{-0.004}
\newcommand{\maticfphwins}{2/6}

\newcommand{\matcnnlstmgldelta}{-0.001}
\newcommand{\matcnnlstmglwins}{3/6}

\newcommand{\matcnnlstmphdelta}{-0.010}
\newcommand{\matcnnlstmphwins}{1/6}

\newcommand{\matcnngldelta}{+0.001}
\newcommand{\matcnnglwins}{2/6}

\newcommand{\matcnnphdelta}{+0.028}
\newcommand{\matcnnphwins}{4/6}

\newcommand{\ridgegl}{0.59}
\newcommand{\ridgeglsd}{0.04}
\newcommand{\ridgeglvert}{0.69}
\newcommand{\ridgeph}{0.28}
\newcommand{\ridgephsd}{0.31}
\newcommand{\ridgephvert}{0.40}
\newcommand{\recglall}{0.615}

\newcommand{\recglvals}{-0.174}

\newcommand{\recphall}{0.648}
\newcommand{\recphbase}{0.630}

\newcommand{\recphvals}{0.631, 0.739}

\newcommand{\ablglapgain}{0.57}
\newcommand{\mskgaingllo}{0.04}
\newcommand{\mskgainglhi}{0.08}

\newcommand{\mskgainphlo}{0.02}
\newcommand{\mskgainphhi}{0.03}

\newcommand{\uvertgllo}{9.9}
\newcommand{\uvertglhi}{13.1}
\newcommand{\upeakgllo}{14.4}
\newcommand{\upeakglhi}{19.6}
\newcommand{\ucopgllo}{7.4}
\newcommand{\ucopglhi}{8.5}

\newcommand{\uvertphlo}{10.7}
\newcommand{\uvertphhi}{11.4}
\newcommand{\upeakphlo}{10.4}
\newcommand{\upeakphhi}{11.0}
\newcommand{\ucopphlo}{5.1}
\newcommand{\ucopphhi}{7.4}

\newcommand{\newtonglntrials}{279}
\newcommand{\newtonglrawforward}{-0.11}

\newcommand{\newtonsglrawforward}{-0.13}
\newcommand{\newtonglrawlateral}{0.42}

\newcommand{\newtonglrawvertical}{0.48}

\newcommand{\newtonsglrawvertical}{0.52}
\newcommand{\newtonglrotforward}{0.29}

\newcommand{\newtonsglrotforward}{0.36}
\newcommand{\newtonglrotlateral}{0.44}

\newcommand{\newtonglrotvertical}{0.48}

\newcommand{\newtonsglrotvertical}{0.52}
\newcommand{\newtonphntrials}{434}
\newcommand{\newtonphrawforward}{0.01}

\newcommand{\newtonsphrawforward}{0.00}
\newcommand{\newtonphrawlateral}{-0.03}

\newcommand{\newtonphrawvertical}{0.21}

\newcommand{\newtonsphrawvertical}{0.25}
\newcommand{\newtonphrotforward}{0.04}

\newcommand{\newtonsphrotforward}{0.06}
\newcommand{\newtonphrotlateral}{-0.08}

\newcommand{\newtonphrotvertical}{0.21}

\newcommand{\newtonsphrotvertical}{0.25}
}

\title{Subject-Held-Out Evaluation of Deep Ground Reaction Force Estimators Across Two Laboratories: Leakage and Cross-Laboratory Transfer}

\author{Hernan~David~Santos,
        Yinong~Chen,
        Yu~Ding,
        Xudong~Zhang,
        and~Jian~Tao%
\ifdefined\cleancopy\else\thanks{Manuscript received \todo{date}.}\fi%
\thanks{This work has been submitted to the IEEE for possible publication. Copyright may be transferred without notice, after which this version may no longer be accessible.}%
\thanks{This work was supported in part by the Ocean Energy Safety Institute Consortium (OESIC) through a grant from the U.S. Department of the Interior, Bureau of Safety and Environmental Enforcement (BSEE), and the U.S. Department of Energy (DOE), and was accomplished under Agreement Number E21AC00000.}%
\thanks{H.~D.~Santos is with the Department of Electrical and Computer Engineering and the Texas A\&M Institute of Data Science, Texas A\&M University, College Station, TX 77843 USA (e-mail: hsantos@tamu.edu).}%
\thanks{Y.~Chen is with the Department of Environmental and Occupational Health, Texas A\&M University, College Station, TX 77843 USA.}%
\thanks{Y.~Ding is with the H.~Milton Stewart School of Industrial and Systems Engineering, Georgia Institute of Technology, Atlanta, GA 30332 USA.}%
\thanks{X.~Zhang is with the Wm Michael Barnes '64 Department of Industrial and Systems Engineering, Texas A\&M University, College Station, TX 77843 USA.}%
\thanks{J.~Tao is with the College of Performance, Visualization and Fine Arts and the Texas A\&M Institute of Data Science, Texas A\&M University, College Station, TX 77843 USA (e-mail: jtao@tamu.edu).}%
\ifdefined\cleancopy\else\thanks{\todo{Author order and affiliations to be confirmed with all co-authors.}}\fi}

\begin{document}
\maketitle
\ifdefined\cleancopy\else
\begin{center}
\fcolorbox{red}{red!4}{\parbox{0.93\columnwidth}{\footnotesize\color{red!80!black}
\textbf{DRAFT STATUS (October 19, 2026), remove before submission.} Pending items are marked \todo{like this} in the text.
(1) \textbf{Venue:} IEEE JBHI (decided Oct 6; TBME is the fallback). All experiments are complete; every GroundLink result was retrained after the unit fix of Oct 8--10.
(2) \textbf{To do, David:} read the eight audit papers known only from their abstracts; make the code repository public before reviewers open it.
(3) \textbf{Done:} author consent form signed by all authors; OESI funding statement (Agreement E21AC00000); IRB approval date and informed consent in Section~III-A.
(4) \textbf{To do, Dr.\ Zhang / Yinong:} confirm the Patient Handling mediolateral plate sign convention (Section~\ref{sec:res_frame}).
(5) \textbf{Page budget:} the funding footnote and Acknowledgment bring the main text to 9 pages (page 9 is charged); with the supplement the total is 15, one over 14.
}}
\end{center}
\fi

\begin{abstract}
Deep networks that estimate ground reaction forces and centers of pressure from motion-capture markers are validated within one laboratory, in a seldom-stated coordinate frame, and sometimes with trial-level splits that put the same person in the training and the test set. We measured what each practice costs, and whether such models transfer between laboratories. Three sequence models (convolutional, convolutional-recurrent, and convolutional-transformer) were evaluated on two independent datasets, locomotion (6 subjects) and occupational patient handling (10 subjects), under one pipeline, one tuning procedure, three seeds, and four protocols: trial-level split, leave-one-subject-out, pooled two-laboratory training with matched held-out subjects, and cross-laboratory transfer. At a fixed configuration, trial-level splits overstated the subject-held-out coefficient of determination by \inflglcnn{} to \inflphcnn{}; on occupational data the overstatement exceeded every difference between architectures. With subjects held out, the three models reached values between \hphcnn{} and \hglcnnlstm{}; the leader differed between datasets, and gaps lay within subject-to-subject variability. Pooling the laboratories changed neither dataset on matched subjects. Transfer between laboratories was below chance in both directions, including the vertical force, which no in-plane rotation can affect. Leaving force targets in the plate frame while inputs were heading-aligned cost \ablglapgain{} of locomotion anteroposterior-force explained variance, enough to make that force look hardest; a Newton's-law correlation detected the mismatch before training. Split protocol and target frame changed reported accuracy more than architecture did. Learned force estimators should be validated with subjects held out, checked for frame consistency before training, and tested across laboratories before deployment claims are made.
\end{abstract}


\section{Introduction}
\label{sec:intro}

\IEEEPARstart{G}{round} reaction forces (GRFs) and centers of pressure (CoP) are the reference inputs for joint-load estimation, gait assessment, and ergonomic risk evaluation, and force plates are the instrument that measures them. Plates are expensive, fixed to a few square meters of floor, and absent from clinics, workplaces, and field studies, so a model that recovers GRF and CoP from motion-capture markers alone would remove the main constraint on where kinetic analysis can be done. A growing body of work trains deep networks for exactly this mapping and reports encouraging accuracy: CNN--LSTM models with \rsq{} near 0.74 for vertical force in running~\cite{chen2025cnnxlstm}, feedforward networks with correlations above 0.9 from inertial data~\cite{mundt2020estimation}, and attention-based models with correlations above 0.96 in cross-dataset tests~\cite{faisal2024multiresunet}.

Two evaluation practices limit what these numbers say about deployment. In an audit of forty learned GRF estimators (Table~\ref{tab:audit}), about half demonstrably hold subjects out, a third do not state the split unit, and one in five quote a within-subject result as the headline, with cross-subject error two to five times larger wherever both are reported~\cite{jiang2020single,saeb2017,karbalaie2026participant}. When trials of the same person appear on both sides of a split, a model can exploit person-specific signatures instead of the kinematic-to-kinetic mapping that a new subject requires~\cite{horst2019explaining,halilaj2018machine}. Second, only one audited paper in five states the coordinate frame of its force targets, and only one in eight tests on a second laboratory or dataset with measured forces~\cite{johnson2021predicting,faisal2024multiresunet,tan2026gaitdynamics}, although every deployment of a virtual force plate is a new laboratory, clinic, or worksite, and a change of laboratory changes the capture system, the marker protocol, the coordinate convention, the tasks, and the population at once.

This paper evaluates three representative sequence models, a convolutional network, a CNN--LSTM, and a CNN--Transformer hybrid, on two datasets collected in different laboratories for different purposes: GroundLink locomotion~\cite{han2023groundlink} and an occupational patient-handling dataset~\cite{chen2025biodynamic}. Everything runs through one pipeline and one set of protocols: trial-level split, leave-one-subject-out (LOSO), pooled two-laboratory training with matched held-out subjects, and cross-laboratory transfer. The contribution is the evaluation and what it exposes, not a new architecture. We make four points.

\textbf{Leakage inflation, measured on two task domains.} With the same configuration on both protocols, trial-level splits overstate subject-held-out \rsq{} by \inflglcnn{} to \inflphcnn{} (Section~\ref{sec:res_leak}). The inflation appears on two datasets that share nothing but a marker subset and for every model, so it is a property of the protocol, not of a dataset; on the occupational data it is larger than any difference between architectures.

\textbf{An unstated target frame can manufacture a channel ordering.} Only eight of forty audited papers state the frame of their force targets. Leaving the targets in the force-plate frame while the inputs are heading-aligned lowers locomotion anteroposterior-force \rsq{} from \ablglcorotap{} to \ablglbaselineap{}, making that channel look intrinsically the hardest; a Newton's-law correlation on the preprocessed data flags the mismatch before training (Section~\ref{sec:res_frame}).

\textbf{No universal architecture, with uncertainty stated.} Tuned with one procedure and evaluated across seeds, the three models reach LOSO \rsq{} between \hphcnn{} and \hglcnnlstm{}, the leader changes between datasets, and between-model gaps lie inside subject-to-subject variability (Section~\ref{sec:res_models}). A memoryless ridge baseline on the same features reaches \ridgegl{} and \ridgeph{}, so the engineered features carry much of the signal and temporal modeling adds the rest.

\textbf{Cross-laboratory transfer fails, and not because of the coordinate frame.} Models trained in one laboratory and tested in the other are far below chance in both directions, and the vertical channel, which no in-plane rotation can touch, fails with the horizontal ones (Section~\ref{sec:res_xlab}). Pooling both laboratories, compared on matched held-out subjects, changes neither dataset. We decompose the gap into four sources, coordinate convention, task distribution, marker placement and hardware, and population, and state which the present design attributes and which it confounds. Laboratory-held-out tests of learned GRF estimation are rare (five of forty audited papers) and disagree on whether the gap is large~\cite{johnson2021predicting} or absent once every source is harmonized through one pipeline~\cite{tan2026gaitdynamics}; it is the level at which a deployment claim has to be tested.

The two datasets are small, 6 and 10 subjects, and differ in every property at once, so the cross-laboratory result is a stress test and not a controlled study; we say so wherever it appears. All code, configurations, per-run hyperparameters, and run outputs are released so that every number in this paper can be regenerated.

\section{Related Work}
\label{sec:related}

\subsection{Learned GRF Estimation}
Estimating kinetics from kinematics has moved from musculoskeletal simulation~\cite{delp2007opensim} to learned regressors on marker, inertial, or video inputs. Reported models span feedforward networks~\cite{mundt2020estimation}, recurrent and convolutional hybrids~\cite{chen2025cnnxlstm,johnson2021predicting,alcantara2022predicting}, attention-based encoders~\cite{faisal2024multiresunet}, and, most recently, models trained on large simulated corpora with zero-shot transfer to measured data~\cite{liu2025imdy}. The GroundLink dataset~\cite{han2023groundlink} paired markers with plates for locomotion and reported a baseline under a subject split. We audited forty trained GRF, moment, or CoP estimators from 2013 to 2026 (Table~\ref{tab:audit}; the supplementary audit table lists each paper, its split, and the source of each classification). Twenty-one demonstrably hold subjects out, twelve do not state the split unit in accessible text, and seven do not hold subjects out, six of them inertial-sensor papers that quote a within-subject number as the headline; eight state the frame of their force targets, and five test on a second laboratory or dataset with measured forces. Earlier reviews tabulated validation along other axes~\cite{gurchiek2019estimating,xiang2022recent,oliveira2026scoping}, and none audited the target frame.

\begin{table}[t]\centering
\caption{Evaluation protocols of forty published learned GRF estimators (each paper, its split, and the source of each classification: Supplementary Table~S11). Held out: test subjects absent from training. Not: trial-level split, or a within-subject result quoted as the headline. n.s.: split unit not stated. Frame: target frame stated. 2nd site: tested on another laboratory or dataset with measured forces.}
\label{tab:audit}\footnotesize\setlength{\tabcolsep}{3.5pt}
\begin{tabular}{lccccc}\toprule
\textbf{Input} & \textbf{Papers} & \textbf{Held out} & \textbf{Not / n.s.} & \textbf{Frame} & \textbf{2nd site} \\ \midrule
Optical markers & 13 & 6 & 0 / 7 & 2 & 1 \\
Inertial sensors & 13 & 6 & 6 / 1 & 3 & 2 \\
Video & 12 & 8 & 1 / 3 & 2 & 1 \\
Multi-study & 2 & 1 & 0 / 1 & 1 & 1 \\
\midrule
All & 40 & 21 & 7 / 12 & 8 & 5 \\
\bottomrule\end{tabular}\end{table}

\subsection{Subject Leakage and Held-Out Evaluation}
That trials of one person should not straddle a split is well known in gait analysis~\cite{halilaj2018machine,horst2019explaining} and in adjacent fields with hierarchical data, where random splits inflate accuracy through temporal or identity leakage~\cite{saeb2017,hammerla2015}. Quantified comparisons of record-wise and subject-wise splitting exist for classification, where the inflation is 5 to 33 accuracy points~\cite{saeb2017,karbalaie2026participant}; for GRF regression the published comparisons are within- versus cross-subject results inside single papers, where the cross-subject error is two to five times larger~\cite{jiang2020single}. What has not been available is one estimator retrained under both splits on the same data, with the gap reported for a continuous kinetic target on more than one task domain, which is what Section~\ref{sec:res_leak} provides.

\subsection{Coordinate Frames and Cross-Site Shift}
Marker trajectories and forces are expressed in laboratory frames that differ between sites, and heading alignment is the standard remedy for the kinematics. That the force vector must be transformed with the same rotation is elementary but easy to omit when kinematics and kinetics are processed by different code paths, and only eight of the forty audited papers state which frame their targets are in, while no best-practice guide treats the choice as an evaluation issue~\cite{halilaj2018machine}; Section~\ref{sec:res_frame} measures what the omission costs. Equivariant and geometric architectures~\cite{bronstein2021geometric,thomas2018tensor,xu2024egno} remove the dependence by construction and are a natural next step, but they do not address task, hardware, or population shift, which Section~\ref{sec:res_xlab} shows to dominate the two-laboratory gap. Larger multi-site corpora such as AddBiomechanics~\cite{werling2023addbiomechanics} make a controlled decomposition of these sources possible and are where this evaluation should be repeated.

\section{Materials and Methods}
\label{sec:methods}

\subsection{Datasets}
\label{sec:data}

\textbf{GroundLink (GL)}~\cite{han2023groundlink} contains locomotion and sport movements (walking, running, jumping, cutting) from 7 subjects captured with a Qualisys system and two AMTI plates, markers at \SI{250}{Hz} and forces at \SI{250}{Hz}, both resampled to \SI{50}{Hz}. GL uses a $Z$-up convention with $F_x$ anteroposterior (AP), $F_y$ mediolateral (ML), $F_z$ vertical. Subject 1 shows a constant per-channel offset on both plates consistent with an un-zeroed plate and is excluded from the main analyses; 6 subjects contribute 279 trials (26 to 57 per subject).

\textbf{Patient Handling (PH)}~\cite{chen2025biodynamic} contains occupational lifting, transfer, and repositioning tasks from 12 subjects captured with a Vicon system and two AMTI plates, markers at \SI{50}{Hz} and forces at \SI{1000}{Hz} resampled to \SI{50}{Hz}, in a $Y$-up OpenSim convention ($v_x$ ML, $v_y$ vertical, $v_z$ AP). Two subjects have their plate-1 and plate-2 channels physically swapped and are excluded from the main analyses; the remaining 10 subjects recorded 460 trials, of which 26 with missing essential markers or more than 10\% missing samples are dropped at loading, leaving 434 (33 to 46 per subject). The Patient Handling data were collected under a protocol approved by the Texas A\&M University Institutional Review Board (STUDY2024-0979, approved October 1, 2024), and all participants gave written informed consent; GroundLink is a public dataset.

Both exclusions are calibration errors with a known repair (a data-driven tare for GL subject 1, a plate swap for the two PH subjects). Section~\ref{sec:res_models} reports the models with the repaired subjects reinstated. Table~\ref{tab:datasets} summarizes the two datasets. They were not chosen as a controlled cross-laboratory pair: they are the two marker-plus-plate datasets available to us that span different task domains, and they differ simultaneously in task, hardware, marker protocol, coordinate convention, calibration, and population.

\begin{table}[t]
\centering
\caption{Dataset characteristics.}
\label{tab:datasets}
\footnotesize
\begin{tabular}{lcc}
\toprule
 & \textbf{GroundLink} & \textbf{Patient Handling} \\
\midrule
Task domain & Locomotion, sport & Occupational handling \\
Subjects used / recorded & 6 / 7 & 10 / 12 \\
Trials used & 279 & 434 \\
Trials per subject & 26--57 & 33--46 \\
Capture system & Qualisys & Vicon \\
Marker protocol & 24 markers & 20 markers \\
Laboratory frame & $Z$-up & $Y$-up (OpenSim) \\
Marker / force rate (Hz) & 250 / 250 & 50 / 1000 \\
Mean vertical force (N) & 249.5 & 386.2 \\
\bottomrule
\end{tabular}
\end{table}

\subsection{Inputs, Targets, and Preprocessing}
\label{sec:prep}

All models receive the 20-marker intersection of the two protocols (bilateral shoulder, elbow, wrist, pelvis, thigh, knee, shank, ankle, heel, toe) and predict ten channels per frame: for each plate, ML force, vertical force, AP force, and the two horizontal CoP coordinates. Preprocessing has five stages, applied identically to both datasets.

\begin{enumerate}
\item \textbf{Pelvis rooting.} Marker positions are expressed relative to the pelvis centroid (mean of the two anterior superior iliac spine markers).
\item \textbf{Heading alignment.} Each trial is rotated about the vertical axis by an angle $\theta$ so that the pelvis anterior direction, the trial-averaged vector from the posterior to the anterior iliac-spine midpoints, points along $+Z$. Anatomy fixes the sign, so there is no ambiguity. The same rotation is applied to the CoP coordinates and to the pelvis-derived center-of-mass (COM) velocity and acceleration features.
\item \textbf{Height scaling.} Positions are divided by subject height.
\item \textbf{Feature engineering.} Positions, first and second finite differences, ten bilateral inter-marker distances, and the COM kinematics give about 216 features per frame.
\item \textbf{Standardization.} Inputs are z-scored; forces are divided by body weight ($F/mg$, with $m$ the body mass in each dataset's metadata; GroundLink forces are released in kilonewtons and are converted to newtons first) and then z-scored per channel with statistics from the training fold only.
\end{enumerate}

\begin{figure*}[t]
\centering
\includegraphics[width=0.95\textwidth]{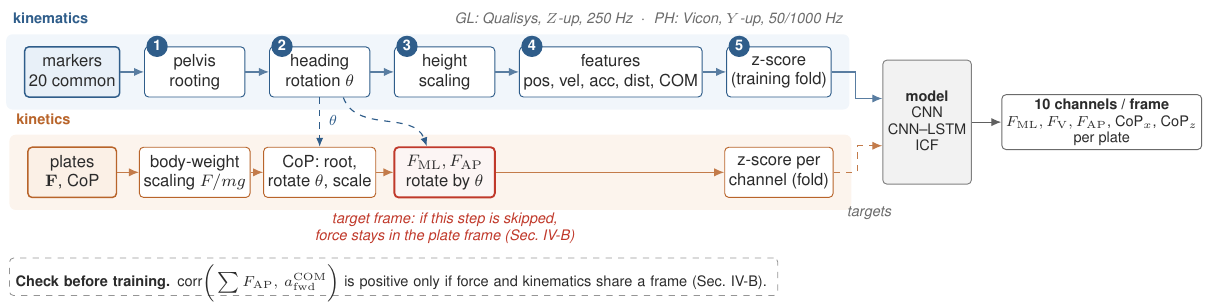}
\caption{Preprocessing pipeline. Kinematic inputs (top) and kinetic targets (bottom) are processed by separate code paths that must share the per-trial heading rotation $\theta$. Applying $\theta$ to the markers, the CoP, and the COM features but not to the horizontal force components (red box) leaves the targets in the plate frame; Section~\ref{sec:res_frame} measures the consequence.}
\label{fig:pipeline}
\end{figure*}

\textbf{Target frame.} Force is a vector. Stage 2 rotates the marker inputs, the CoP, and the COM features by $\theta$; the question is whether the ML and AP force targets are rotated too. If they are left in the force-plate frame, the model is asked to predict a horizontal force whose axes are rotated by a per-trial angle $\theta$, discarded after preprocessing, relative to the frame of its inputs. We call these \emph{plate-frame targets}. The alternative, \emph{co-rotated targets}, rotates the horizontal force components with the same $\theta$,
\begin{equation}
\begin{bmatrix} F'_{\mathrm{ML}} \\ F'_{\mathrm{AP}} \end{bmatrix}
= \begin{bmatrix} \cos\theta & \sin\theta \\ -\sin\theta & \cos\theta \end{bmatrix}
\begin{bmatrix} F_{x} \\ F_{z} \end{bmatrix},
\label{eq:corot}
\end{equation}
with $(F_x, F_z)$ the plate-frame horizontal components in each dataset's axis assignment. Vertical force is invariant to the rotation. A second choice with a similar flavor is where the input standardization statistics come from: per trial, which discards each trial's absolute velocity and acceleration scale, or from the training fold. Section~\ref{sec:res_frame} reports the two choices as a $2\times2$ factorial at fixed hyperparameters; all other results use co-rotated targets and per-fold standardization, and because the standardization choice affects cross-laboratory transfer in its own right, Section~\ref{sec:res_xlab} reports transfer under both.

\textbf{Frame-consistency diagnostic.} Newton's second law gives $\sum_i \mathbf{F}_i = m\,\mathbf{a}_{\mathrm{COM}}$ for the whole body when the plates carry all external force. If the summed horizontal GRF and the COM acceleration are expressed in the same frame, their components are positively correlated along each axis; if the frames differ by an unknown rotation, the correlation is destroyed. We compute, per trial, the Pearson correlation between the summed AP force and the forward COM acceleration, and between the summed ML force and the lateral COM acceleration, using the pipeline's own features and targets; the vertical pair, unaffected by any in-plane rotation, is the control. The check requires no trained model and runs in seconds.

\textbf{Augmentation.} Training trials are mirrored left--right (markers and plate assignments swapped, ML components negated), doubling the training set without altering geometry.

\subsection{Models}
\label{sec:models}

Three sequence-to-sequence models map the $T\times D$ feature sequence to $T\times10$ outputs. \textbf{CNN}: three residual 1-D convolutional blocks (kernels 3, 5, 7; batch normalization; GELU) and a linear head. \textbf{CNN--LSTM}: the same encoder followed by a bidirectional LSTM and a linear head~\cite{hochreiter1997long,schuster1997bidirectional}, representative of the hybrids common in GRF estimation~\cite{chen2025cnnxlstm}. \textbf{CNN--Transformer (ICF)}: the same encoder, a linear projection to $d_{\mathrm{model}}$, broadcast addition of a linear embedding of subject mass and height, learnable positional embeddings, $L$ pre-LayerNorm Transformer layers with 8 heads~\cite{xiong2020layer}, and a linear head. No dataset or laboratory indicator is given to any model. Hyperparameters are selected per model and per dataset (Section~\ref{sec:hpo}); Table~S1 in the supplement lists the configuration behind every reported result; parameter counts range from 0.9M to 4.1M.

All models minimize a peak-weighted mean squared error on the standardized targets, $\mathcal{L} = \frac{1}{TC}\sum_{t,c} (1+\lambda\tilde{y}_{t,c})(\hat{y}_{t,c}-y_{t,c})^2$ with $\tilde{y}$ the per-batch min--max normalized target and $\lambda=2$, so that loaded frames weigh up to three times more than unloaded ones. Accuracy is the coefficient of determination per channel on the held-out subject's frames, $R^2_c = 1-\sum_t(\hat{y}_{t,c}-y_{t,c})^2/\sum_t(y_{t,c}-\bar{y}_c)^2$, averaged over the ten channels; negative values mean worse than predicting the test mean. We also report mean absolute error in body weights where a comparison requires it.

\subsection{Evaluation Protocols}
\label{sec:protocols}

\textbf{Trial-level split.} An 80/10/10 random split of trials with subjects mixed across the three parts, the protocol most common in the literature, run with the same configuration as LOSO so that the two protocols differ only in what is held out.

\textbf{Leave-one-subject-out (LOSO).} Every trial of one subject forms the test set, the remaining subjects form training and validation (grouped by subject, so no subject straddles the two), and all normalization statistics are fit on the training portion; 6 folds for GL and 10 for PH.

\textbf{Pooled training with matched comparison.} One model is trained on both datasets; each of six folds holds out one GL subject and one PH subject and evaluates each on its own laboratory's trials. Because only six of the ten PH subjects are held out, the pooled result is compared with single-laboratory LOSO on the same six subjects, as a paired per-subject difference.

\textbf{Cross-laboratory transfer.} Each LOSO model is also evaluated on every trial of the other laboratory, with no subject of that laboratory in training. Input standardization uses the training laboratory's statistics, which is the deployment condition.

\subsection{Hyperparameter Selection, Training, and Statistics}
\label{sec:hpo}

Each model is tuned separately on each dataset with the same procedure: Optuna tree-structured Parzen estimation with Hyperband pruning~\cite{akiba2019optuna}, 60 trials of at most 60 epochs, objective the \rsq{} on one held-out proxy subject per dataset (gl\_5, ph\_12), fixed before the search. The proxy was not an easy subject: under the final three-seed LOSO, gl\_5 is tied with gl\_4 as the hardest of the six GL subjects and ph\_12 is the third hardest of ten (Fig.~\ref{fig:subjects}), so tuning on it does not favor the reported means. The ICF search includes depth $L\in\{2,3,4\}$ and width $d_{\mathrm{model}}\in\{32,\dots,256\}$; a separate depth-by-width grid confirmed that two layers match or exceed four at every width on GL, and its best configuration is kept as a second ICF row. Selected configurations are then run under full LOSO, under the trial-level split, and, for the frame study, replaced by one fixed ICF configuration so that only the target convention varies; all are listed in Table~S1 of the supplement. The GroundLink searches predate our fix of a unit error (kilonewtons read as body weights); their configurations were kept and every GroundLink model was retrained.

Training uses AdamW~\cite{loshchilov2019decoupled}, cosine annealing with warm restarts, gradient clipping at norm 1, up to 120 epochs with early stopping on validation loss (patience 15), on one consumer GPU. Headline LOSO numbers are the mean over held-out subjects of the per-subject \rsq{}, repeated for three seeds (42, 43, 44) and reported as mean $\pm$ SD across seeds; every other table uses seed 42. Models are compared with the Wilcoxon signed-rank test on paired per-subject \rsq{}; with 6 and 10 folds the smallest attainable two-sided $p$ is 0.031 and 0.002, so borderline values are reported as such rather than dichotomized.

\section{Results}
\label{sec:results}

\subsection{Trial-Level Splits Overstate Subject-Held-Out Accuracy}
\label{sec:res_leak}

\begin{figure}[t]
\centering
\includegraphics[width=\columnwidth]{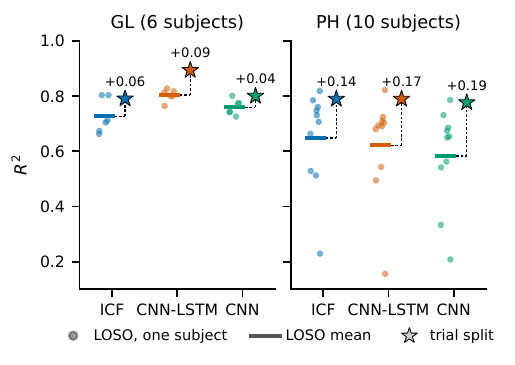}
\caption{Trial-level split versus LOSO, same configuration per model; all scores are means over three training seeds. Numbers give the gap between the trial-split score and the LOSO mean, which is positive for every model.}
\label{fig:inflation}
\end{figure}

Fig.~\ref{fig:inflation} and the last two columns of Table~\ref{tab:headline} compare each model under a trial-level split and under LOSO with the same configuration and pipeline. \trialpending{} For the ICF the trial split reaches \rsq{} \tsglicf{} on GL and \tsphicf{} on PH; LOSO reaches \hglicf{} and \hphicf{}. The difference, \inflglicf{} and \inflphicf{}, is the accuracy attributable to seeing the test subject's other trials during training. It appears on two datasets that share only a marker subset and for all three models (\inflglcnnlstm{} and \inflglcnn{} on GL, \inflphcnnlstm{} and \inflphcnn{} on PH). On PH it is larger than any between-architecture difference reported below; on GL it is of the same order as those differences.

\subsection{The Target Frame Sets the Channel Ordering}
\label{sec:res_frame}

\begin{figure}[t]
\centering
\includegraphics[width=\columnwidth]{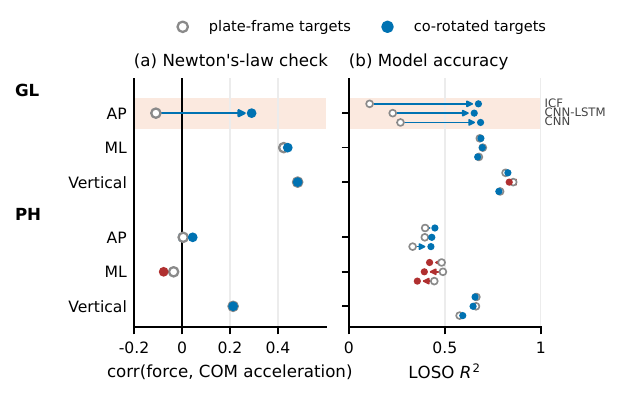}
\caption{Plate-frame versus co-rotated force targets (per-trial standardization). (a) Mean per-trial correlation of summed force with center-of-mass acceleration, computed before training. (b) LOSO \rsq{}, mean of the two plates; top to bottom in each row: ICF, CNN--LSTM, CNN. Only GL AP force moves, in both panels and for every model; red marks a decrease.}
\label{fig:frame}
\end{figure}

Before comparing models we settle whether the horizontal force targets are rotated with the heading-aligned inputs (co-rotated) or left in the force-plate frame (Section~\ref{sec:prep}). Fig.~\ref{fig:frame} compares the two with per-trial standardization, for the ICF at one fixed configuration ($d_{\mathrm{model}}=64$, $L=2$) and the other two models at their tuned ones; Table~S3 of the supplement adds per-fold standardization. On GL, where the heading rotation averages about $138^\circ$, the Newton's-law check flags the mismatch before training: summed AP force correlates with forward COM acceleration at \newtonglrawforward{} with plate-frame targets and \newtonglrotforward{} with co-rotated ones, while the rotation-invariant vertical control stays at \newtonglrawvertical{} (Fig.~\ref{fig:frame}a). The models agree: co-rotation lifts ICF AP-force \rsq{} from \ablglbaselineap{} to \ablglcorotap{}, and that of the CNN--LSTM and CNN by \ablcnnlstmglapgain{} and \ablcnnglapgain{} (Table~S3); with per-fold standardization the ICF ten-channel mean rises from \ablglbaseliner{} to \ablglbothr{}, a paired gain of \ablpairedgldelta{} on \ablpairedglwins{} subjects ($p=\ablpairedglp{}$). On PH, whose trials already face the aligned heading, neither panel moves much; its gain (\ablphbaseliner{} to \ablphbothr{}) comes from per-fold standardization, which restores each trial's absolute velocity and acceleration scale. Co-rotation lowers PH ML-force \rsq{} for all three models (\ablphmldrop{} to \ablcnnlstmphmldrop{}), and the PH lateral correlation is weakly negative (\newtonphrotlateral{}), which suggests a mirrored ML sign convention.

With plate-frame targets AP force is the hardest channel on GL for every model (Fig.~\ref{fig:frame}b), the ``vertical and CoP easiest, AP hardest'' hierarchy; with co-rotated targets it rises to the level of ML force. What survives is weaker and symmetric: ML and AP force are both harder than vertical force and CoP$_x$ on both datasets, by \SI{0.1}{} to \SI{0.3}{} of \rsq{}. All results that follow use co-rotated targets with per-fold standardization.

\subsection{No Architecture Dominates}
\label{sec:res_models}

\begin{figure}[t]
\centering
\includegraphics[width=\columnwidth]{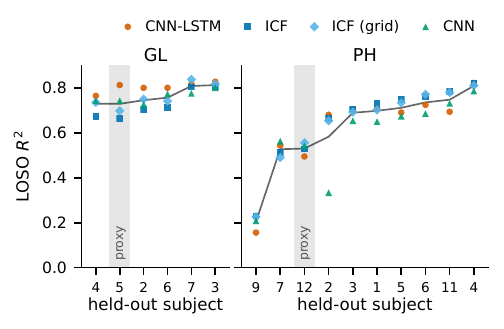}
\caption{Per-subject LOSO \rsq{} (mean of three seeds) for each held-out subject, sorted from hardest to easiest. The shaded column is the proxy subject used as the tuning objective; on both datasets it sits in the harder half.}
\label{fig:subjects}
\end{figure}

Table~\ref{tab:headline} reports the three models, each tuned with the same procedure, under LOSO with three seeds. \seedpending{} On GL the CNN--LSTM leads (\hglcnnlstm{}), followed by the ICF grid configuration (\hglicfg{}), the CNN (\hglcnn{}), and the ICF configuration chosen by the search (\hglicf{}); on PH the ICF leads (\hphicf{}), followed by the ICF grid configuration (\hphicfg{}), the CNN--LSTM (\hphcnnlstm{}), and the CNN (\hphcnn{}). The ranking reverses between datasets. Within a dataset the between-model differences (\pglcnnlstmicfd{} on GL, $p=\pglcnnlstmicf{}$; \pphcnnlstmicfd{} on PH, $p=\pphcnnlstmicf{}$) exceed the seed-to-seed spread of any one model (SD \hglcnnlstmsd{} to \hphcnnsd{} across three seeds), so training noise does not explain them; but the differences are of the same size as the effect of which proxy subject the search happened to tune on, smaller than the subject-to-subject spread (Fig.~\ref{fig:subjects}; Supplement Tables~S4 and~S5), and the leader is a different model on the two datasets. All three models generalize to new subjects; none is the model to prefer.

\begin{table}[t]
\centering
\caption{Subject-held-out \rsq{}, each model tuned with the same procedure. LOSO: mean over held-out subjects, $\pm$ the \seedsd{}; ML and AP are the mean of the two plates. Trial: the same configuration under a trial-level 80/10/10 split, mean of three training seeds; $\Delta$ is the leakage inflation (trial minus LOSO).}
\label{tab:headline}
\footnotesize
\setlength{\tabcolsep}{2.8pt}
\renewcommand{\arraystretch}{1.15}
\begin{tabular}{llccccc}
\toprule
\textbf{Data} & \textbf{Model} & \textbf{LOSO \rsq} & \textbf{ML} & \textbf{AP} & \textbf{Trial} & $\boldsymbol{\Delta}$ \\
\midrule
GL & CNN--LSTM & \hglcnnlstm{} $\pm$ \hglcnnlstmsd{} & \hglcnnlstmml{} & \hglcnnlstmap{} & \tsglcnnlstm{} & \inflglcnnlstm{} \\
GL & ICF (grid, $d$=128) & \hglicfg{} $\pm$ \hglicfgsd{} & \hglicfgml{} & \hglicfgap{} & -- & -- \\
GL & CNN & \hglcnn{} $\pm$ \hglcnnsd{} & \hglcnnml{} & \hglcnnap{} & \tsglcnn{} & \inflglcnn{} \\
GL & ICF (search, $d$=96) & \hglicf{} $\pm$ \hglicfsd{} & \hglicfml{} & \hglicfap{} & \tsglicf{} & \inflglicf{} \\
GL & Ridge (memoryless) & \ridgegl{} $\pm$ \ridgeglsd{} & -- & -- & -- & -- \\
\midrule
PH & ICF (search, $d$=96) & \hphicf{} $\pm$ \hphicfsd{} & \hphicfml{} & \hphicfap{} & \tsphicf{} & \inflphicf{} \\
PH & ICF (grid, $d$=128) & \hphicfg{} $\pm$ \hphicfgsd{} & \hphicfgml{} & \hphicfgap{} & -- & -- \\
PH & CNN--LSTM & \hphcnnlstm{} $\pm$ \hphcnnlstmsd{} & \hphcnnlstmml{} & \hphcnnlstmap{} & \tsphcnnlstm{} & \inflphcnnlstm{} \\
PH & CNN & \hphcnn{} $\pm$ \hphcnnsd{} & \hphcnnml{} & \hphcnnap{} & \tsphcnn{} & \inflphcnn{} \\
PH & Ridge (memoryless) & \ridgeph{} $\pm$ \ridgephsd{} & -- & -- & -- & -- \\
\bottomrule
\end{tabular}
\end{table}

\textbf{Linear baseline.} A per-channel ridge regression on the same 216 features, folds, and normalization, with no temporal context, reaches \rsq{} \ridgegl{} on GL and \ridgeph{} on PH (vertical force: \ridgeglvert{} and \ridgephvert{}). The features carry much of the signal; temporal modeling adds about 0.2 of \rsq{} on GL and 0.4 on PH.

\textbf{Physical units and unloaded CoP.} Per trial, LOSO vertical-force RMSE is \uvertgllo{} to \uvertglhi{}\,\%BW on GL and \uvertphlo{} to \uvertphhi{}\,\%BW on PH, the peak is missed by \upeakgllo{} to \upeakglhi{} and \upeakphlo{} to \upeakphhi{}\,\%BW, and CoP on a loaded plate by \ucopgllo{} to \ucopglhi{} and \ucopphlo{} to \ucopphhi{}\,cm. Scoring CoP only while its plate carries over 5\% of body weight (CoP is undefined on an unloaded plate, yet the \rsq{} above includes those frames) raises LOSO \rsq{} by \mskgaingllo{} to \mskgainglhi{} on GL and \mskgainphlo{} to \mskgainphhi{} on PH and changes no leader or conclusion.

\textbf{Exclusion sensitivity.} Reinstating the excluded subjects with their calibration errors repaired changes the conclusions only where expected. On PH the two swapped-plate subjects score \recphvals{} after the swap is undone and the twelve-subject mean is \recphall{} against \recphbase{} for ten. On GL the tared subject 1 scores \recglvals{}, well below every other subject, and drags the seven-subject mean to \recglall{}; its offset is not a pure tare, and we keep it excluded while reporting the number.

\subsection{Pooling Two Laboratories}
\label{sec:res_pool}

Table~\ref{tab:pooled} reports the models trained on both laboratories under dual-holdout cross-validation, with the matched comparison against single-laboratory LOSO on the same held-out subjects. Pooling neither helps nor hurts: on the same held-out subjects the change is \maticfgldelta{} (ICF) and \matcnnlstmgldelta{} (CNN--LSTM) on GL and \maticfphdelta{} and \matcnnlstmphdelta{} on PH, with no paired test below $p=0.3$. The matched comparison matters: the six PH subjects that fall in the pooled folds are easier than the ten-subject average, so an unmatched comparison would show a spurious PH gain. A pooled model matches each single-laboratory model on its own data. With six matched subjects this is an observation, not a transfer law.

\begin{table}[t]
\centering
\caption{Pooled two-laboratory training. Dual-holdout \rsq{} on the held-out GL and PH subject of each fold (mean $\pm$ SD over six folds), and the paired difference $\Delta$ against single-laboratory LOSO on the same subjects, with the number of subjects improved by pooling in parentheses.}
\label{tab:pooled}
\scriptsize
\setlength{\tabcolsep}{3.5pt}
\begin{tabular}{lcccc}
\toprule
 & \multicolumn{2}{c}{\textbf{GL held-out}} & \multicolumn{2}{c}{\textbf{PH held-out}} \\
\cmidrule(lr){2-3}\cmidrule(lr){4-5}
\textbf{Model} & Pooled & $\Delta$ (improved) & Pooled & $\Delta$ (improved) \\
\midrule
ICF & \combicfgl{} $\pm$ \combicfglsd{} & \maticfgldelta{} (\maticfglwins{}) & \combicfph{} $\pm$ \combicfphsd{} & \maticfphdelta{} (\maticfphwins{}) \\
CNN--LSTM & \combcnnlstmgl{} $\pm$ \combcnnlstmglsd{} & \matcnnlstmgldelta{} (\matcnnlstmglwins{}) & \combcnnlstmph{} $\pm$ \combcnnlstmphsd{} & \matcnnlstmphdelta{} (\matcnnlstmphwins{}) \\
CNN & \combcnngl{} $\pm$ \combcnnglsd{} & \matcnngldelta{} (\matcnnglwins{}) & \combcnnph{} $\pm$ \combcnnphsd{} & \matcnnphdelta{} (\matcnnphwins{}) \\
\bottomrule
\end{tabular}
\end{table}

\subsection{Cross-Laboratory Transfer Fails, and Not Because of the Frame}
\label{sec:res_xlab}

Table~\ref{tab:xlab} reports models trained entirely in one laboratory and tested in the other, under both input standardizations. Every entry is negative. With per-trial standardization the mean \rsq{} ranges from \xlsubglphicf{} to \xlsubphglcnnlstm{}; with per-fold standardization fit on the training laboratory, the deployment condition, it is worse still, and the CNN trained on PH collapses on GL (\xlphglcnn{}) because the GL inputs fall far outside the PH training statistics. Per-trial standardization is an implicit, instance-level adaptation that the per-fold variant removes; the choice that is correct within a laboratory is the harsher cross-laboratory test.

\begin{table}[t]
\centering
\caption{Cross-laboratory transfer \rsq{} (mean over source-laboratory LOSO models, evaluated on every trial of the other laboratory), co-rotated targets, under per-trial and per-fold input standardization. Vertical-force \rsq{} is shown for the per-fold case; it is invariant to any in-plane rotation.}
\label{tab:xlab}
\footnotesize
\setlength{\tabcolsep}{4pt}
\begin{tabular}{llccc}
\toprule
\textbf{Direction} & \textbf{Model} & \textbf{Per-trial} & \textbf{Per-fold} & \textbf{Vertical (per-fold)} \\
\midrule
GL $\to$ PH & ICF & \xlsubglphicf{} & \xlglphicf{} & \xlglphicfvert{} \\
GL $\to$ PH & CNN--LSTM & \xlsubglphcnnlstm{} & \xlglphcnnlstm{} & \xlglphcnnlstmvert{} \\
GL $\to$ PH & CNN & \xlsubglphcnn{} & \xlglphcnn{} & \xlglphcnnvert{} \\
PH $\to$ GL & ICF & \xlsubphglicf{} & \xlphglicf{} & \xlphglicfvert{} \\
PH $\to$ GL & CNN--LSTM & \xlsubphglcnnlstm{} & \xlphglcnnlstm{} & \xlphglcnnlstmvert{} \\
PH $\to$ GL & CNN & \xlsubphglcnn{} & \xlphglcnn{} & \xlphglcnnvert{} \\
\bottomrule
\end{tabular}
\end{table}

\begin{figure}[t]
\centering
\includegraphics[width=\columnwidth]{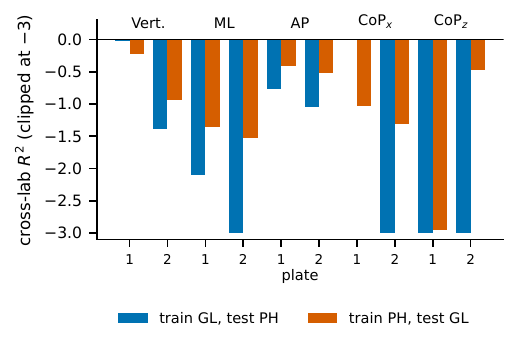}
\caption{Per-channel cross-laboratory \rsq{} (ICF, per-fold standardization), clipped at $-3$ for display. Vertical force fails with the horizontal channels.}
\label{fig:xlab}
\end{figure}

Two facts locate the cause. First, co-rotation, which changed within-laboratory AP accuracy by half a point of \rsq{}, does not rescue transfer: with force and kinematics in one consistent heading frame in both laboratories, the rotational part of the coordinate convention is no longer a candidate explanation. Second, the vertical channel fails as well (\xlglphicfvert{} and \xlphglicfvert{} for the ICF; no model exceeds \xlphglcnnlstmvert{}), and no rotation about the vertical axis can affect it. What remains is what the two laboratories do not share: task (gait and jumping against lifting and transferring, with an 11-fold difference in body-weight-normalized AP-force variance), marker placement and hardware, plate placement and calibration (a 14-fold difference in CoP range), and population. Section~\ref{sec:disc_xlab} sets out which of these a two-laboratory design can and cannot separate. Fig.~\ref{fig:xlab} shows the per-channel transfer scores.

\section{Discussion}
\label{sec:discussion}

\subsection{What the Protocol Measures}
The leakage inflation of \inflglcnn{} to \inflphcnn{} \rsq{} is as large as the architecture effects in this paper on GL, larger than all of them on PH, and larger than most reported between-method differences in the literature. A trial-level split answers the question ``how well does this model predict more trials of people it has already seen,'' which is not the question a virtual force plate is built to answer. LOSO is the minimum protocol for a subject-generalization claim; the datasets here are small enough that it costs nothing, and on larger corpora it costs one grouped split. We recommend that GRF-estimation papers state the split unit explicitly and report LOSO whenever the model is intended for new people.

\subsection{Architecture Choice Is Second Order}
Once the pipeline is consistent and each model is tuned by the same procedure, the three architectures are within subject-to-subject variability of each other, and the leader changes between datasets. We do not think the reversal reflects a deep inductive-bias story; with 6 and 10 subjects it reflects which subjects are hard and which proxy fold the search saw. The target-frame result makes the same point from the other side: one preprocessing decision moved locomotion AP-force \rsq{} by half a point, ten times any between-model gap. The channel ordering that survives, horizontal force harder than vertical and CoP$_x$ by 0.1 to 0.3, is what acceleration-dependent targets estimated from position data should cost; the locomotion-specific AP deficit belongs to the plate-frame convention. The practical implication for a virtual force plate is that data, protocol, and preprocessing decisions dominate model selection at this scale. A three-seed ablation of the pipeline (Supplement Table~S10) makes the same point once more: no preprocessing step is universally right. Heading alignment earns half a point of AP-force \rsq{} on locomotion and costs 0.02 on patient handling, where the subject faces one way and the alignment moves signal between the horizontal channels; pelvis-relative marker coordinates cost 0.07 on locomotion, because per-plate targets need the body's position over the plates, and help by 0.02 on patient handling; velocity and acceleration features matter on patient handling and not on locomotion; augmentation, relative-position features and the peak-weighted loss are within seed noise. The steps that help are the ones with a physical reason on that task, and the reason changes with the task.

\subsection{Sources of the Cross-Laboratory Gap}
\label{sec:disc_xlab}
A model moved between laboratories meets four changes at once. \emph{Coordinate convention}: the axes, and signs, in which markers and forces are expressed. \emph{Task distribution}: which movements, and hence which force waveforms, are present. \emph{Marker placement and hardware}: protocol, marker count, capture and plate systems, calibration. \emph{Population}: anthropometrics and movement strategy. The present design attributes one term and confounds three. Heading alignment applied consistently to kinematics and kinetics removes the rotational part of the convention term by construction, and Section~\ref{sec:res_xlab} shows that transfer still fails, so rotation is not the binding constraint between these two laboratories. The Newton's-law check also hints at a residual sign convention in the PH mediolateral channel; a sign is a convention term too, and it is one that the same diagnostic can catch before training. The vertical-channel failure and the 11-fold difference in AP-force variance point to task and calibration, but task, hardware, and population cannot be separated with one laboratory pair. Separating them needs data in which one term changes and the others do not: the same laboratory and protocol recording different tasks isolates task; the same laboratory recording the same tasks with a fuller marker set isolates placement; a multi-laboratory corpus such as AddBiomechanics~\cite{werling2023addbiomechanics} with study and laboratory identifiers allows the laboratory to be the held-out unit. That is the evaluation this paper argues for and does not yet perform.

The pooling result adds one constraint. Training on both laboratories neither helps nor hurts either dataset on matched subjects. A pooled model does not learn a laboratory-invariant representation that improves on each laboratory; it fits two conditional ones side by side. Equivariant architectures~\cite{bronstein2021geometric,xu2024egno} would remove the rotational convention term more elegantly than heading alignment but, on this evidence, would not by themselves close a gap that rotation does not explain.

\subsection{Lessons}
\label{sec:lessons}
Three checks came out of this study, and each is cheap. First, run the Newton's-law correlation on the preprocessed tensors before training; it takes seconds and it distinguishes a frame mismatch from a hard channel. Second, tabulate the configuration behind every reported number; hyperparameter searches run per dataset and per protocol select different models, and a Methods section that describes one configuration where several were used will be caught. Third, fit input statistics on the training fold, not per trial; per-trial standardization hides scale information within a laboratory and acts as an undeclared adaptation across laboratories. A fourth concerns the leakage measurement itself: when the hyperparameter search is run on the trial-level split, it rewards capacity that memorizes subjects, selects models several times larger, and roughly doubles the apparent inflation; Table~\ref{tab:headline} therefore holds the configuration fixed across protocols. We learned all of these the hard way: an earlier version of this study used plate-frame targets, per-trial standardization, and split-tuned models, and the audit prompted by its review produced the frame study of Section~\ref{sec:res_frame}.

\subsection{Limitations}
Six and ten subjects give six and ten paired observations per comparison, so the model rankings and the pooling effect are indicative. The two datasets differ in everything at once, so the cross-laboratory result is a stress test, not a decomposition. The hyperparameter search used one proxy subject per dataset for cost, and the GL ranking shows that the choice matters at this scale. Trial counts on GL are unbalanced across subjects. The frame study uses one seed per arm.

\section{Conclusion}
\label{sec:conclusion}

We evaluated three deep GRF and CoP estimators on locomotion and occupational data from two laboratories under one pipeline and three protocols. Trial-level splits overstate subject-held-out \rsq{} by \inflglcnn{} to \inflphcnn{}. Expressing the force targets in the force-plate frame while the kinematic inputs are heading-aligned costs half a point of \rsq{} on locomotion anteroposterior force and is enough to make AP force look intrinsically hardest; a Newton's-law check exposes the mismatch before training. With frame-consistent targets and every model tuned by the same procedure, all three architectures generalize to new subjects and none dominates. Pooling the laboratories neither helps nor hurts on matched subjects; transfer between laboratories fails in both directions for reasons the coordinate frame does not explain, since the rotation-invariant vertical channel fails with the rest. The results argue for three practices in learned biomechanics: hold out the subject, check that vector targets share the inputs' frame before training, and treat the laboratory, not the trial or the subject, as the unit that a deployment claim must survive. Code, configurations, and all run outputs are available at \url{https://github.com/hsantos10/Metahuman_Attention}.

\section*{Acknowledgment}
Computational resources were provided by Texas A\&M High Performance Research Computing. The views and conclusions contained in this document are those of the authors and should not be interpreted as representing the opinions or policies of the U.S. Government. Mention of trade names or commercial products does not constitute their endorsement by the U.S. Government.\todo{Acknowledge the Patient Handling data-collection team?}

\bibliographystyle{IEEEtran}
\bibliography{main}

\clearpage
\setcounter{section}{0}\setcounter{table}{0}\setcounter{figure}{0}
\renewcommand{\thetable}{S\arabic{table}}\renewcommand{\thefigure}{S\arabic{figure}}\renewcommand{\thesection}{S\arabic{section}}
\begin{center}{\Large Supplementary Material}\end{center}

\section{Configurations}
Table~\ref{tab:configs} lists the configuration behind every result in the main text.
\begin{table*}[!h]
\centering
\caption{Configuration behind each reported result. FFN is the feed-forward width for ICF and the LSTM hidden size for CNN--LSTM; $k$ is the convolution kernel; parameter counts are obtained by instantiating each configuration with the 216-feature input.}
\label{tab:configs}
\footnotesize
\setlength{\tabcolsep}{6pt}
\begin{tabular}{llccccc}
\toprule
\textbf{Result} & \textbf{Model} & $d_{\mathrm{model}}$ & $L$ & FFN & Filters ($k$) & Params \\
\midrule
\multicolumn{7}{l}{\emph{Tuned per model and dataset: LOSO and trial split (main-text Tables~II and III, Figs.~2 and~4)}} \\
GL & ICF (search) & 96 & 2 & 1024 & 128 & 0.93M \\
GL, PH & ICF (grid) & 128 & 2 & 512 & 256 & 1.61M \\
GL & CNN--LSTM & -- & 2 & 512 & 256 (7) & 4.07M \\
GL & CNN & -- & 3 & -- & 512 (7) & 2.62M \\
PH & ICF (search) & 96 & 2 & 512 & 256 & 1.42M \\
PH & CNN--LSTM & -- & 1 & 256 & 256 (7) & 0.92M \\
PH & CNN & -- & 3 & -- & 256 (15) & 1.82M \\
\multicolumn{7}{l}{\emph{Fixed configuration: frame study, pooling, transfer (main-text Fig.~3, Tables~IV and~V)}} \\
GL, PH, pooled & ICF & 64 & 2 & 512 & 256 & 1.25M \\
GL / PH & CNN--LSTM & -- & 1 & 512 / 256 & 256 (7) & 1.97M / 0.92M \\
GL / PH & CNN & -- & 3 & -- & 512 (3) / 256 (7) & 1.12M / 2.62M \\
\bottomrule
\end{tabular}
\end{table*}

\section{Target frame: diagnostic and factorial}
Table~\ref{tab:newton} gives the Newton's-law frame check of the main text in full: the mean per-trial Pearson correlation between summed bilateral force and pelvis (COM) acceleration along the same axis, for plate-frame and co-rotated force targets, computed on the pipeline's own features and targets. The vertical pair is invariant to any in-plane rotation and serves as the control. With plate-frame targets the model is asked to predict horizontal force in a frame its inputs cannot see, and the damage scales with the heading angle, which is why it is largest on GroundLink (mean rotation about $138^\circ$) and small on Patient Handling, whose trials already face the aligned heading. The weakly negative lateral correlation on Patient Handling (\newtonphrawlateral{} to \newtonphrotlateral{}, more negative with smoothing) suggests that its mediolateral force sign is mirrored relative to the kinematic $X$ axis; a consistent sign is learnable within one laboratory and does not affect the LOSO results, but it is a convention term across laboratories. Table~\ref{tab:ablation} gives the full $2\times2$ factorial of target convention and input standardization for the ICF at the fixed configuration and for the CNN--LSTM and CNN at their tuned configurations.

\begin{table}[H]
\centering
\caption{Newton's-law frame check: mean per-trial Pearson correlation between summed bilateral force and pelvis (COM) acceleration along the same axis, for plate-frame and co-rotated force targets. Vertical is invariant to the rotation and serves as the control.}
\label{tab:newton}
\footnotesize
\setlength{\tabcolsep}{5pt}
\renewcommand{\arraystretch}{1.15}
\begin{tabular}{llccc}
\toprule
\textbf{Dataset} & \textbf{Targets} & \textbf{AP} & \textbf{ML} & \textbf{Vertical} \\
\midrule
GL (\newtonglntrials{} trials) & plate frame & \newtonglrawforward{} & \newtonglrawlateral{} & \newtonglrawvertical{} \\
 & co-rotated & \newtonglrotforward{} & \newtonglrotlateral{} & \newtonglrotvertical{} \\
PH (\newtonphntrials{} trials) & plate frame & \newtonphrawforward{} & \newtonphrawlateral{} & \newtonphrawvertical{} \\
 & co-rotated & \newtonphrotforward{} & \newtonphrotlateral{} & \newtonphrotvertical{} \\
\bottomrule
\end{tabular}
\end{table}

\begin{table}[H]
\centering
\caption{Frame study: LOSO \rsq{} (mean over subjects) under the four combinations of target convention and input standardization, for the ICF at one fixed configuration ($d_{\mathrm{model}}=64$, $L=2$) and for the CNN--LSTM and CNN at their tuned configurations (seed 42). All main-text results use co-rotated targets with per-fold standardization.}
\label{tab:ablation}
\footnotesize
\setlength{\tabcolsep}{2.3pt}
\renewcommand{\arraystretch}{1.15}
\begin{tabular}{lllcccccc}
\toprule
\textbf{Data} & \textbf{Targets} & \textbf{Input std.} & \textbf{Mean} & \textbf{Vert.} & \textbf{ML} & \textbf{AP} & \textbf{CoP$_x$} & \textbf{CoP$_z$} \\
\midrule
\multicolumn{9}{l}{\textit{ICF}} \\
GL & plate & trial & \ablglbaseliner{} & \ablglbaselinevert{} & \ablglbaselineml{} & \ablglbaselineap{} & \ablglbaselinepx{} & \ablglbaselinepz{} \\
GL & co-rot. & trial & \ablglcorotr{} & \ablglcorotvert{} & \ablglcorotml{} & \ablglcorotap{} & \ablglcorotpx{} & \ablglcorotpz{} \\
GL & plate & fold & \ablglperfoldr{} & \ablglperfoldvert{} & \ablglperfoldml{} & \ablglperfoldap{} & \ablglperfoldpx{} & \ablglperfoldpz{} \\
GL & co-rot. & fold & \ablglbothr{} & \ablglbothvert{} & \ablglbothml{} & \ablglbothap{} & \ablglbothpx{} & \ablglbothpz{} \\
\cmidrule(lr){1-9}
PH & plate & trial & \ablphbaseliner{} & \ablphbaselinevert{} & \ablphbaselineml{} & \ablphbaselineap{} & \ablphbaselinepx{} & \ablphbaselinepz{} \\
PH & co-rot. & trial & \ablphcorotr{} & \ablphcorotvert{} & \ablphcorotml{} & \ablphcorotap{} & \ablphcorotpx{} & \ablphcorotpz{} \\
PH & plate & fold & \ablphperfoldr{} & \ablphperfoldvert{} & \ablphperfoldml{} & \ablphperfoldap{} & \ablphperfoldpx{} & \ablphperfoldpz{} \\
PH & co-rot. & fold & \ablphbothr{} & \ablphbothvert{} & \ablphbothml{} & \ablphbothap{} & \ablphbothpx{} & \ablphbothpz{} \\
\midrule
\multicolumn{9}{l}{\textit{CNN--LSTM}} \\
GL & plate & trial & \ablcnnlstmglbaseliner{} & \ablcnnlstmglbaselinevert{} & \ablcnnlstmglbaselineml{} & \ablcnnlstmglbaselineap{} & \ablcnnlstmglbaselinepx{} & \ablcnnlstmglbaselinepz{} \\
GL & co-rot. & trial & \ablcnnlstmglcorotr{} & \ablcnnlstmglcorotvert{} & \ablcnnlstmglcorotml{} & \ablcnnlstmglcorotap{} & \ablcnnlstmglcorotpx{} & \ablcnnlstmglcorotpz{} \\
GL & plate & fold & \ablcnnlstmglperfoldr{} & \ablcnnlstmglperfoldvert{} & \ablcnnlstmglperfoldml{} & \ablcnnlstmglperfoldap{} & \ablcnnlstmglperfoldpx{} & \ablcnnlstmglperfoldpz{} \\
GL & co-rot. & fold & \ablcnnlstmglbothr{} & \ablcnnlstmglbothvert{} & \ablcnnlstmglbothml{} & \ablcnnlstmglbothap{} & \ablcnnlstmglbothpx{} & \ablcnnlstmglbothpz{} \\
\cmidrule(lr){1-9}
PH & plate & trial & \ablcnnlstmphbaseliner{} & \ablcnnlstmphbaselinevert{} & \ablcnnlstmphbaselineml{} & \ablcnnlstmphbaselineap{} & \ablcnnlstmphbaselinepx{} & \ablcnnlstmphbaselinepz{} \\
PH & co-rot. & trial & \ablcnnlstmphcorotr{} & \ablcnnlstmphcorotvert{} & \ablcnnlstmphcorotml{} & \ablcnnlstmphcorotap{} & \ablcnnlstmphcorotpx{} & \ablcnnlstmphcorotpz{} \\
PH & plate & fold & \ablcnnlstmphperfoldr{} & \ablcnnlstmphperfoldvert{} & \ablcnnlstmphperfoldml{} & \ablcnnlstmphperfoldap{} & \ablcnnlstmphperfoldpx{} & \ablcnnlstmphperfoldpz{} \\
PH & co-rot. & fold & \ablcnnlstmphbothr{} & \ablcnnlstmphbothvert{} & \ablcnnlstmphbothml{} & \ablcnnlstmphbothap{} & \ablcnnlstmphbothpx{} & \ablcnnlstmphbothpz{} \\
\midrule
\multicolumn{9}{l}{\textit{CNN}} \\
GL & plate & trial & \ablcnnglbaseliner{} & \ablcnnglbaselinevert{} & \ablcnnglbaselineml{} & \ablcnnglbaselineap{} & \ablcnnglbaselinepx{} & \ablcnnglbaselinepz{} \\
GL & co-rot. & trial & \ablcnnglcorotr{} & \ablcnnglcorotvert{} & \ablcnnglcorotml{} & \ablcnnglcorotap{} & \ablcnnglcorotpx{} & \ablcnnglcorotpz{} \\
GL & plate & fold & \ablcnnglperfoldr{} & \ablcnnglperfoldvert{} & \ablcnnglperfoldml{} & \ablcnnglperfoldap{} & \ablcnnglperfoldpx{} & \ablcnnglperfoldpz{} \\
GL & co-rot. & fold & \ablcnnglbothr{} & \ablcnnglbothvert{} & \ablcnnglbothml{} & \ablcnnglbothap{} & \ablcnnglbothpx{} & \ablcnnglbothpz{} \\
\cmidrule(lr){1-9}
PH & plate & trial & \ablcnnphbaseliner{} & \ablcnnphbaselinevert{} & \ablcnnphbaselineml{} & \ablcnnphbaselineap{} & \ablcnnphbaselinepx{} & \ablcnnphbaselinepz{} \\
PH & co-rot. & trial & \ablcnnphcorotr{} & \ablcnnphcorotvert{} & \ablcnnphcorotml{} & \ablcnnphcorotap{} & \ablcnnphcorotpx{} & \ablcnnphcorotpz{} \\
PH & plate & fold & \ablcnnphperfoldr{} & \ablcnnphperfoldvert{} & \ablcnnphperfoldml{} & \ablcnnphperfoldap{} & \ablcnnphperfoldpx{} & \ablcnnphperfoldpz{} \\
PH & co-rot. & fold & \ablcnnphbothr{} & \ablcnnphbothvert{} & \ablcnnphbothml{} & \ablcnnphbothap{} & \ablcnnphbothpx{} & \ablcnnphbothpz{} \\
\bottomrule
\end{tabular}
\end{table}

\section{Per-subject and per-fold results}
Tables~\ref{tab:s1_gl}--\ref{tab:s4} list every number behind the means reported in the main text. Subject identifiers follow the released run outputs.

\begin{table}[H]\centering\caption{Per-subject LOSO \rsq{} on GL, corrected pipeline, re-tuned configurations, seed 42.}\label{tab:s1_gl}\scriptsize
\begin{tabular}{lcccc}\toprule
\textbf{Subject} & \textbf{CNN-LSTM} & \textbf{ICF} & \textbf{ICF-grid} & \textbf{CNN} \\ \midrule
gl\_2 & 0.783 & 0.710 & 0.759 & 0.725 \\
gl\_3 & 0.830 & 0.787 & 0.818 & 0.809 \\
gl\_4 & 0.734 & 0.679 & 0.731 & 0.749 \\
gl\_5 & 0.812 & 0.714 & 0.735 & 0.747 \\
gl\_6 & 0.811 & 0.708 & 0.745 & 0.757 \\
gl\_7 & 0.836 & 0.822 & 0.839 & 0.775 \\
\textbf{mean} & 0.801 & 0.737 & 0.771 & 0.760 \\
\bottomrule\end{tabular}\end{table}

\begin{table}[H]\centering\caption{Per-subject LOSO \rsq{} on PH, corrected pipeline, re-tuned configurations, seed 42.}\label{tab:s1_ph}\scriptsize
\begin{tabular}{lcccc}\toprule
\textbf{Subject} & \textbf{CNN-LSTM} & \textbf{ICF} & \textbf{ICF-grid} & \textbf{CNN} \\ \midrule
ph\_1 & 0.707 & 0.730 & 0.701 & 0.680 \\
ph\_11 & 0.729 & 0.799 & 0.779 & 0.765 \\
ph\_12 & 0.488 & 0.544 & 0.596 & 0.507 \\
ph\_2 & 0.669 & 0.664 & 0.683 & 0.301 \\
ph\_3 & 0.705 & 0.697 & 0.696 & 0.612 \\
ph\_4 & 0.802 & 0.819 & 0.825 & 0.782 \\
ph\_5 & 0.671 & 0.744 & 0.745 & 0.690 \\
ph\_6 & 0.711 & 0.752 & 0.761 & 0.654 \\
ph\_7 & 0.572 & 0.488 & 0.444 & 0.582 \\
ph\_9 & 0.147 & 0.272 & 0.216 & 0.195 \\
\textbf{mean} & 0.620 & 0.651 & 0.645 & 0.577 \\
\bottomrule\end{tabular}\end{table}

\begin{table}[H]\centering\caption{Fold-mean LOSO \rsq{} per seed on GL, corrected pipeline, all folds, three seeds.}\label{tab:s2_gl}\scriptsize
\begin{tabular}{lccccc}\toprule
\textbf{Config} & \textbf{seed 42} & \textbf{seed 43} & \textbf{seed 44} & \textbf{mean} & \textbf{SD} \\ \midrule
CNN-LSTM-hpo & 0.801 & 0.804 & 0.807 & 0.804 & 0.003 \\
CNN-hpo & 0.760 & 0.770 & 0.750 & 0.760 & 0.010 \\
ICF-grid128 & 0.771 & 0.769 & 0.750 & 0.763 & 0.012 \\
ICF-hpo & 0.737 & 0.711 & 0.733 & 0.727 & 0.014 \\
\bottomrule\end{tabular}\end{table}

\begin{table}[H]\centering\caption{Fold-mean LOSO \rsq{} per seed on PH, corrected pipeline, all folds, three seeds.}\label{tab:s2_ph}\scriptsize
\begin{tabular}{lccccc}\toprule
\textbf{Config} & \textbf{seed 42} & \textbf{seed 43} & \textbf{seed 44} & \textbf{mean} & \textbf{SD} \\ \midrule
CNN-LSTM-hpo & 0.620 & 0.616 & 0.629 & 0.622 & 0.007 \\
CNN-hpo & 0.577 & 0.579 & 0.592 & 0.583 & 0.008 \\
ICF-grid128 & 0.645 & 0.638 & 0.642 & 0.642 & 0.003 \\
ICF-hpo & 0.651 & 0.632 & 0.662 & 0.648 & 0.015 \\
\bottomrule\end{tabular}\end{table}

\begin{table}[H]\centering\caption{Pooled dual-holdout CV per fold, corrected pipeline.}\label{tab:s3}\scriptsize
\begin{tabular}{llllcc}\toprule
\textbf{Model} & \textbf{Fold} & \textbf{GL held out} & \textbf{PH held out} & \textbf{GL \rsq} & \textbf{PH \rsq} \\ \midrule
ICF & 1 & gl\_2 & ph\_1 & 0.725 & 0.714 \\
ICF & 2 & gl\_3 & ph\_2 & 0.854 & 0.633 \\
ICF & 3 & gl\_4 & ph\_3 & 0.767 & 0.706 \\
ICF & 4 & gl\_5 & ph\_4 & 0.701 & 0.810 \\
ICF & 5 & gl\_6 & ph\_5 & 0.796 & 0.692 \\
ICF & 6 & gl\_7 & ph\_6 & 0.860 & 0.733 \\
CNN-LSTM & 1 & gl\_2 & ph\_1 & 0.829 & 0.717 \\
CNN-LSTM & 2 & gl\_3 & ph\_2 & 0.857 & 0.640 \\
CNN-LSTM & 3 & gl\_4 & ph\_3 & 0.807 & 0.692 \\
CNN-LSTM & 4 & gl\_5 & ph\_4 & 0.858 & 0.840 \\
CNN-LSTM & 5 & gl\_6 & ph\_5 & 0.835 & 0.676 \\
CNN-LSTM & 6 & gl\_7 & ph\_6 & 0.860 & 0.746 \\
CNN & 1 & gl\_2 & ph\_1 & 0.723 & 0.692 \\
CNN & 2 & gl\_3 & ph\_2 & 0.770 & 0.573 \\
CNN & 3 & gl\_4 & ph\_3 & 0.745 & 0.628 \\
CNN & 4 & gl\_5 & ph\_4 & 0.736 & 0.806 \\
CNN & 5 & gl\_6 & ph\_5 & 0.758 & 0.641 \\
CNN & 6 & gl\_7 & ph\_6 & 0.815 & 0.709 \\
\bottomrule\end{tabular}\end{table}

\begin{table}[H]\centering\caption{Cross-laboratory transfer per source-laboratory LOSO model (ICF, corrected pipeline, per-fold input standardization). Own = held-out subject of the training laboratory; Other = all trials of the other laboratory.}\label{tab:s4}\scriptsize
\begin{tabular}{llcc}\toprule
\textbf{Direction} & \textbf{Held-out subject} & \textbf{Own \rsq} & \textbf{Other-lab \rsq} \\ \midrule
GL $\to$ PH & gl\_2 & 0.759 & -3.322 \\
GL $\to$ PH & gl\_3 & 0.851 & -6.594 \\
GL $\to$ PH & gl\_4 & 0.756 & -3.851 \\
GL $\to$ PH & gl\_5 & 0.722 & -1.926 \\
GL $\to$ PH & gl\_6 & 0.808 & -3.018 \\
GL $\to$ PH & gl\_7 & 0.856 & -3.648 \\
PH $\to$ GL & ph\_1 & 0.725 & -0.802 \\
PH $\to$ GL & ph\_2 & 0.609 & -1.136 \\
PH $\to$ GL & ph\_3 & 0.702 & -1.108 \\
PH $\to$ GL & ph\_4 & 0.821 & -1.164 \\
PH $\to$ GL & ph\_5 & 0.725 & -1.201 \\
PH $\to$ GL & ph\_6 & 0.754 & -1.032 \\
PH $\to$ GL & ph\_7 & 0.551 & -1.234 \\
PH $\to$ GL & ph\_9 & 0.162 & -1.256 \\
PH $\to$ GL & ph\_11 & 0.769 & -0.901 \\
PH $\to$ GL & ph\_12 & 0.538 & -0.924 \\
\bottomrule\end{tabular}\end{table}

\begin{table*}[t]\centering\caption{Preprocessing and training-choice ablation: ICF at the fixed configuration ($d_{\mathrm{model}}=64$, $L=2$), corrected pipeline, full LOSO, one factor removed per row, three training seeds (42, 43, 44). \rsq{}: mean over seeds of the fold-mean LOSO \rsq{} $\pm$ SD across seeds. $\Delta$: paired difference from the full pipeline on the same seeds. ML and AP: mean horizontal-force \rsq{} over plates and seeds. Mass and height inputs, per-fold input standardization, body-weight target normalization, centre-of-mass kinematics, and force co-rotation (wherever a heading exists) are kept in every arm.}\label{tab:s5}\scriptsize
\begin{tabular}{lcccccccc}\toprule
\textbf{Arm} & \textbf{GL \rsq} & \textbf{GL $\Delta$} & \textbf{GL ML} & \textbf{GL AP} & \textbf{PH \rsq} & \textbf{PH $\Delta$} & \textbf{PH ML} & \textbf{PH AP} \\ \midrule
full pipeline & 0.740 $\pm$ 0.014 & -- & 0.739 & 0.701 & 0.623 $\pm$ 0.003 & -- & 0.405 & 0.431 \\
no pelvis-relative markers & 0.802 $\pm$ 0.002 & +0.062 & 0.747 & 0.728 & 0.603 $\pm$ 0.013 & -0.020 & 0.408 & 0.360 \\
no heading alignment (targets in plate frame) & 0.665 $\pm$ 0.013 & -0.076 & 0.727 & 0.299 & 0.645 $\pm$ 0.009 & +0.022 & 0.546 & 0.447 \\
no height scaling & 0.747 $\pm$ 0.014 & +0.006 & 0.739 & 0.708 & 0.622 $\pm$ 0.000 & -0.001 & 0.416 & 0.411 \\
none of the three above & 0.608 $\pm$ 0.011 & -0.132 & 0.693 & 0.190 & 0.652 $\pm$ 0.007 & +0.029 & 0.464 & 0.363 \\
no velocity/acceleration features & 0.728 $\pm$ 0.009 & -0.012 & 0.731 & 0.649 & 0.573 $\pm$ 0.007 & -0.049 & 0.336 & 0.371 \\
no relative-position features & 0.752 $\pm$ 0.009 & +0.012 & 0.742 & 0.729 & 0.641 $\pm$ 0.009 & +0.018 & 0.426 & 0.460 \\
no bilateral flip augmentation & 0.736 $\pm$ 0.004 & -0.004 & 0.730 & 0.660 & 0.627 $\pm$ 0.009 & +0.005 & 0.418 & 0.420 \\
plain MSE (no peak weight) & 0.753 $\pm$ 0.013 & +0.012 & 0.752 & 0.719 & 0.629 $\pm$ 0.005 & +0.006 & 0.417 & 0.442 \\
\bottomrule\end{tabular}\end{table*}

\section{Preprocessing and training-choice ablation}
Table~\ref{tab:s5} removes one factor at a time from the pipeline at the fixed ICF configuration used in the main-text frame study, with three training seeds per arm, so that the contribution of each preprocessing and training choice is visible against seed-to-seed noise (SD 0.003 to 0.014 of the fold mean). Three results are larger than that noise on every seed. First, heading alignment is task-dependent: on GroundLink it is what makes anteroposterior force learnable (AP \rsq{} 0.65 with, 0.24 without; overall $-0.06$ without), while on Patient Handling, where the subject faces the bed throughout, removing it \emph{raises} \rsq{} by $0.02$, entirely on the mediolateral channel (0.41 to 0.55). Second, expressing markers relative to the pelvis costs $0.07$ on GroundLink (16 of 18 subject--seed pairs worse with it): the targets are per plate, so the model needs the body's position over the plates, which pelvis-relative coordinates remove; on Patient Handling, with both feet on the plates throughout, the same step helps by $0.02$. Height scaling costs $0.05$ on GroundLink and is neutral on Patient Handling. Third, the velocity and acceleration features matter on Patient Handling ($-0.05$ without, 28 of 30 pairs worse) and are neutral on GroundLink. Relative-position features, bilateral flip augmentation and the peak-weighted loss are within $\pm0.02$ on both datasets, the peak weight costing GroundLink about $0.02$. The pattern matches the main text's model comparison: there is no universal preprocessing either. Heading alignment is right for a task in which people walk in different directions and wrong for one in which they face one way, and pelvis-relative coordinates are wrong whenever the target is per plate. The full pipeline used in the main text is therefore not the best configuration for GroundLink; the frame and leakage results do not depend on that, since every comparison holds the pipeline fixed, but a virtual force plate built for one task should choose these steps for that task. A mass and height ``metadata off'' arm is not defined under body-weight target normalization, which needs the mass, and was not run.

\section{Frame-consistency diagnostic, smoothed accelerations}
Finite-difference acceleration at \SI{50}{Hz} is noisy, which lowers all correlations in the main-text Newton's-law table without changing their pattern. With the acceleration low-pass filtered by a five-frame moving average (about \SI{4.4}{Hz} cut-off, in the range conventionally used for center-of-mass kinematics), the forward correlation on GroundLink is \newtonsglrawforward{} with plate-frame targets and \newtonsglrotforward{} with co-rotated targets (a five-frame moving average; longer windows raise the co-rotated value further, to \SI{0.41}{} at nine frames), and the vertical control is \newtonsglrawvertical{} and \newtonsglrotvertical{}; on Patient Handling the corresponding values are \newtonsphrawforward{}, \newtonsphrotforward{}, \newtonsphrawvertical{}, and \newtonsphrotvertical{}.

\section{Excluded subjects}
GroundLink subject 1 shows a constant per-channel offset on both plates consistent with an un-zeroed plate; a data-driven tare (subtracting the per-channel mean over unloaded frames) brings the subject's LOSO \rsq{} from below zero to \recglvals{}, still far below the other subjects, so it remains excluded from the main analyses. Patient Handling subjects 13 and 14 have their plate-1 and plate-2 channels swapped; undoing the swap gives \recphvals{}, in line with the other subjects, and the twelve-subject mean is \recphall{}.

\onecolumn
\section{Evaluation-protocol audit}
Table~\ref{tab:audit_full} lists the papers behind the audit summary table of the main text (Section~II-A): learned estimators of ground reaction force, moment, or center of pressure from optical markers, inertial sensors or insoles, video or markerless pose, and multi-study corpora, published 2013--2026, found through publisher databases and citation chasing from the references of this paper and of three earlier reviews. For each paper we recorded the unit of the headline data split, whether test subjects were absent from training, whether the coordinate frame of the force targets is stated, whether a second laboratory or dataset with measured forces was tested, and the headline result as reported. Classifications were made from the paper's own Methods where the full text was accessible and from the abstract otherwise (marked low confidence); an unstated split is recorded as such and never counted as trial-level. Physics-based methods without training, estimators whose targets are joint moments only, and replications are shown for context and excluded from the counts. The source rows, with a supporting quotation for every split classification, are released with the code.

\begin{center}\scriptsize\setlength{\tabcolsep}{3pt}\renewcommand{\arraystretch}{1.1}
\begin{longtable}{p{2.3cm}p{3.6cm}p{0.9cm}p{2.6cm}p{1.1cm}p{0.8cm}p{1.0cm}p{3.6cm}}
\caption{Evaluation-protocol audit of learned ground-reaction-force, moment and center-of-pressure estimators (2013--2026). Split: unit of the headline split (both = within-subject and cross-subject results reported). Held out: subjects absent from training. Frame: coordinate frame of the force targets stated. Second site: test on a second laboratory or dataset with measured forces. $\dagger$ = not a trained estimator or targets other than GRF/CoP; shown for context, excluded from the counts in the main text. Confidence low = classification from an abstract only.}\label{tab:audit_full}\\
\toprule \textbf{Study (venue)} & \textbf{Input} & \textbf{$n$} & \textbf{Split} & \textbf{Held out} & \textbf{Frame} & \textbf{Second site} & \textbf{Headline result} \\ \midrule
\endfirsthead
\toprule \textbf{Study (venue)} & \textbf{Input} & \textbf{$n$} & \textbf{Split} & \textbf{Held out} & \textbf{Frame} & \textbf{Second site} & \textbf{Headline result} \\ \midrule
\endhead
\midrule \multicolumn{8}{r}{\textit{continued}} \\
\endfoot
\bottomrule
\endlastfoot
\multicolumn{8}{l}{\textbf{Optical markers and marker-derived kinematics}} \\
Oh 2013 (J Biomech)\,(low) & Marker-derived segment kinematics (ANN for double support) & not stated & not stated: not stated (10-fold CV; fold unit not g… & not stated & no & no & R 0.991 vertical, 0.985 AP, 0.918 ML \\
Johnson 2018 (Med Biol Eng Comput)\,(low) & Optical marker trajectories (PLS) & not stated & not stated: not stated & not stated & no & no & mean r 0.980 GRF / 0.914 GRM (11 PLS variants) \\
Johnson 2019 (IEEE TBME)\,(low) & Optical markers encoded as images (CNN) & not stated & not stated: not stated & not stated & no & no & r 0.988 GRF / 0.972 GRM; rRMSE 4.3\% / 7.0\% \\
Mundt 2018$^\dagger$ (ISBS Proc) & Marker-derived joint angles (LSTM) & 12 & subject: subject-level (unknown subjects) + with… & mixed & no & no & qualitative: higher accuracy for known than unknown subjects \\
Mundt 2019 (Med Biol Eng Comput)\,(low) & Joint angles during cutting (FFNN); marker-based implied, n… & 55 & not stated: not stated & not stated & no & no & not retrieved (paywalled) \\
Komaris 2019 (IEEE Access) & Shank marker-cluster accelerations (IMU proxy; ANN) & 28 & subject: subject-level (16/6/6 random by subject… & yes & no & no & test RMSE 0.134 BW vertical, 0.041 AP, 0.042 ML \\
Han 2023 (GroundLink) (SIGGRAPH Asia) & 96 markers $\rightarrow$ SMPL-X joint rotations + pelvis (1D CNN) & 7 & subject: subject-level (s007 held out; rotated i… & yes & yes & no & vGRF MSE (BW-norm) 0.11 L / 0.06 R avg (as reproduced in Le 2025 Tabl… \\
Faisal 2024 (Neural Comput Appl) & 5 foot markers (attention MultiResUNet) & 66 (50 +… & within-dataset 5-fold CV: within-dataset 5-fold CV (subject-wise… & yes & no & yes & cross-dataset r 0.96 GRF / 0.86 GRM; within-B GRF r $>$ 0.98 \\
Ozates 2025 (Ann Biomed Eng)\,(low) & Marker-derived joint angles (1D CNN) & 754 (622… & not stated: not stated & not stated & no & no & CP nRMSE $<$ 20.1\%, PCC $>$ 0.84; TD nRMSE $<$ 12.7\%, PCC $>$ 0.94 \\
Guan 2025 (Med Eng Phys) & OpenSim joint angles/moments (LSTM) & not stated & subject: subject-level (LOSO) & yes & no & no & LSTM r 0.961, nRMSE 13.8\% (vertical, left foot); all directions r 0.6… \\
Le 2025 (Phys-GRD) (HSI 2025 / arXiv) & SMPL pose from GroundLink markers (GroundLinkNet + physics… & 7 & subject: subject-level (one participant test; ro… & yes & yes & no & vGRF MSE 0.09 L / 0.06 R vs 0.11 / 0.06 GroundLinkNet (18\% avg improv… \\
Aasmann 2025 (IEEE EMBC)\,(low) & Marker trajectories (RNN, real-time) & 5 (real-t… & not stated: not stated & not stated & no & no & r\_GRF 0.96, r\_CoP 0.81 (real-time, 5 subjects) \\
Meng 2026 (Front Bioeng Biotechn…) & 39 Plug-in Gait markers as graph nodes (ST-GCN + transfer l… & 90 & subject: subject-level (LOSO over 30 patients; h… & yes & no & no & LOSO: vertical r 0.984, rRMSE 5.24\%; AP r 0.956; ML r 0.912 \\
Zaccardi 2026 (Gait \& Posture)\,(low) & Marker trajectories vs joint angles; marker sparsity (Trans… & not stated & not stated: not stated (training-set mean-waveform… & not stated & no & no & RMSE $<$ 10\% BW, R2 $>$ 0.9 all models; only Transformer beat mean-wavefo… \\
\multicolumn{8}{l}{\textbf{Inertial sensors and insoles}} \\
Guo 2017 (Sensors) & 3 accelerometers (L5, C7, forehead; NARMAX) & 9 & trial/clip: trial-level (temporal half/half within… & no & no & no & relative RMSE 3.8\% (L5, controlled), 5.0\% (free); rho 0.993 \\
Wouda 2018 (Front Physiol) & 3 Xsens IMUs (ANN) & 8 & both: both: trial-level speed hold-out within… & mixed & no & no & single-subject rho $>$ 0.99, RMSE $<$ 0.27 BW; LOSO rho $>$ 0.9 in 7/8 subj… \\
Dorschky 2020 (Front Bioeng Biotechn…) & 4 IMUs + simulated IMU from optimal-control simulation (CNN) & 10 & subject: subject-level fixed hold-out (S01-S03) & yes & no & no & GRF RMSE ~3\% BW AP, ~10\% BW vertical; r 0.971 / 0.980 \\
Mundt 2020$^\dagger$ (Front Bioeng Biotechn…) & 5 IMUs + IMU simulated from marker database (ANN) & 23 measur… & subject: subject-level (LOSO) & yes & no & no & moments nRMSE 13.0\% (measured) / 11.6\% (sim + measured) \\
Jiang 2020 (Sensors) & Single IMU (foot/shank/thigh; random forest) & 9 & both: both: intra-participant temporal half s… & mixed & no & no & shank: intra r 1.00, RMSE 0.02 BW, NRMSE 1.7\%; LOSO r 0.97, RMSE 0.10… \\
Refai 2020$^\dagger$ (Sensors) & Pelvis IMU, error-state EKF (non-ML) & 3 & n/a: n/a (no training; per-subject calibrati… &  & yes & no & shear RMS error 5.2\% BW; vertical r 0.8; shear r 0.5 \\
Johnson 2021 (IEEE TBME) & 5 accelerometers at test; training on accelerations simulat… & test 5 (t… & study: study/lab-level (train UWA archive, tes… & yes & yes & yes & best r Fz 0.97, Fy 0.96, Fx 0.87; mean GRF r 0.89, GRM r 0.65 (sidest… \\
Cerfoglio 2021$^\dagger$ (Sensors) & IMUs (NARX network) & 11 & subject: subject-level fixed hold-out (7 train /… & yes & not assessed & no & GRF RMSE 0.02 N/kg; knee moment RMSE 0.04 N.m/kg \\
Alcantara 2022 (PeerJ) & 3 biaxial accelerometers (RNN) & 19 & subject: subject-level (LOSO); plus single-subje… & yes & yes & no & RMSE 0.16 BW, rRMSE 6.4\% \\
Tan 2024 (IEEE TBME) & 8 IMUs; self-supervised transformer pretrained on MoVi + AM… & 19 / 17 /… & subject: subject-level 5-fold CV & yes & no & no & treadmill walking vGRF r 0.97, RMSE 0.07 N/kg; overground r 0.94 \\
Chen 2025 (Sensors) & Xsens IMU-derived joint angles (CNN-xLSTM) & 12 & 10-fold CV on shuffled p: 10-fold CV on shuffled pool of 530 stan… & not stated & no & no & R2 0.909, MAPE 2.18, rMSE 0.061 \\
Song 2025 (IEEE TBME) & 3 IMUs (SVD embedding regression vs LSTM) & 44 & three scenarios: LOSO (': three scenarios: LOSO ('others'), withi… & mixed & no & no & vGRF RMSE 0.130 BW, rRMSE 4.2\% (personal scenario) \\
Li 2025 (Sensors) & Single shank IMU (wavelet NN vs FFNN) & 15 & subject: subject-level fixed hold-out (1 RFS + 1… & yes & no & no & waveform NRMSE 4.6-6.8\% (WNN); peak MAPE 1.4-6.0\% \\
Olugbon 2025 (arXiv (preprint)) & 96-sensor pressure insole + midfoot IMU (ANN/RF/BiLSTM) & 8 & both: both: intra-participant 70/30 (trial-le… & mixed & no & no & intra RMSE 0.024 BW, NRMSE 1.79\%; LOSO RMSE 0.044 BW, NRMSE 3.22\% \\
Nagashio 2026 (AHFE Int.) & Shoe IMU + 8 plantar pressure sensors per shoe (Transformer) & 12 & subject: subject-level fixed (8 train / 2 val /… & yes & yes & no & NRMSE 5.91\%, R2 0.842 (Fx 0.655, Fy 0.891, Fz 0.981) \\
Lin 2026 (IEEE TNSRE) & 13 IMUs; optimised 4- and 2-IMU subsets (CNN-BiLSTM) & 61 PD + 6… & both: both: intra-subject 80/20 (trial-level)… & mixed & no & yes & inter-subject PD R2 0.91, MAE 3.99 \%BW; HC R2 0.93; intra R2 0.98 \\
Cordero-Sanchez 2025$^\dagger$ (J Appl Biomech)\,(low) & Inertial mocap (training) $\rightarrow$ optical kinematics from public… & not stated & not stated: not stated (cross-source by design) & not stated & no & yes & not retrieved (abstract only) \\
\multicolumn{8}{l}{\textbf{Video, markerless pose and body-model motion}} \\
Scott 2020 (ECCV) & 2D keypoints (OpenPose) from monocular video & 10 & subject: subject-level (LOSO) & yes & partial & no & pressure MAE 8.95 kPa; CoP error in mm (figures only) \\
Rempe 2020 (ECCV) & Monocular video; contact net trained on rendered Mixamo moc… & 13 synthe… & trial/clip: trial/clip-level (80/10/10 by motions p… & no & partial & pose only & contact acc 94.6\% synthetic / 92.7\% real; GRF only vs literature BW r… \\
Shimada 2020 (PhysCap) (ACM TOG (SIGGRAPH Asi…) & Monocular video $\rightarrow$ contact net $\rightarrow$ PD + contact-force QP & not stated & not stated: not stated (no force split; pose protoc… & not stated & no & pose only & MPJPE 68.9 mm; GRF 'reasonable force range' (Fig. 9) \\
Shimada 2021 (Neural PhysCap) (ACM TOG (SIGGRAPH)) & Monocular video $\rightarrow$ GRFNet/torque nets $\rightarrow$ differentiable dyn… & not stated & study: study/dataset-level (DeepCap held out) & yes & no & pose only & MPJPE 63.6 mm on unseen DeepCap; vGRF 'reasonable range' \\
Morris 2021 (ISBS Proc)\,(low) & 2D keypoints from monocular video (estimator not named) & 14 & not stated: not stated & not stated & no & no & r 0.936-0.954, nRMSE 11.1-13.1\% (AP, vertical); ML poorer \\
Louis 2022 (ACM Multimedia) & 2D/3D keypoints from multi-view video treated as monocular… & 8 + 5 & subject: subject-level (ForcePose 6 train / 2 va… & yes & partial & no & ForcePose RMSE 83.2 N; Parkour 83.8 / 85.8 N per sole \\
Mundt 2022 (Sensors) & OpenPose 2D keypoints from synthesised and real video & 14 & subject: subject-level (LOSO) & yes & no & no & r $>$ 0.954, nRMSE 13.3\% (synthesised); real video r 0.943, nRMSE 13.55\% \\
Mundt 2023 (Sensors) & 2D pose from 25-fps video: AlphaPose vs BlazePose vs OpenPo… & 10 (14 re… & subject: subject-level (LOSO; 40 ANNs) & yes & no & no & running sagittal: vertical/AP r $>$ 0.937, nRMSE $<$ 0.20; ML r -0.07 to… \\
Uhlrich 2023 (OpenCap) (PLoS Comput Biol) & 2+ smartphone videos $\rightarrow$ 2D pose $\rightarrow$ LSTM marker augmentation… & 10 valida… & new validation cohort vs: new validation cohort vs legacy trainin… & yes & no & no & GRF MAE 6.2\% BW averaged (11.4\% vertical, 3.5\% AP, 3.8\% ML); KAM MAE… \\
Verheul 2024$^\dagger$ (J Biomech) & OpenCap v0.2 (3 iPads) $\rightarrow$ Newtonian segmental acceleration… & 15 & n/a: n/a (no training) &  & no & yes & Bland-Altman bias $<$ 5\% for 6 variables, 5-15\% for 17; peak landing wo… \\
Katsu 2024 (GRF-MV)$^\dagger$ (BMVC Workshop ANIMA) & Monocular video $\rightarrow$ HybrIK-X SMPL-X $\rightarrow$ physics contact optim… & 10 motion… & n/a: n/a (no task-specific training) &  & no & yes & normalised MSE 0.003-0.150 per motion \\
Zhang 2024 (PhysPT) (CVPR) & Monocular video via SMPL estimates; self-supervised on AMASS & not stated & study: study/dataset-level (AMASS train; other… & yes & yes & pose only & acc error 2.5 mm/frame2; foot sliding 2.6 mm; no force-plate comparis… \\
Le 2024 (OSDCap) (NeurIPS) & Monocular video (TRACE pose) $\rightarrow$ meta-PD $\rightarrow$ simulation $\rightarrow$ le… & H3.6M S1,… & subject: subject-level (standard protocols) & yes & yes & pose only & MPJPE 54.8 mm; foot skating 15.2\% (vs 37.0\%) \\
Liu 2025 (ImDy) (ICLR) & SMPL-format keypoints + velocities; supervision from IsaacG… & not state… & trial/clip: trial/clip-level (27,501 train / 3,055… & not stated & no & yes & GroundLink zero-shot per-foot GRF 0.986 L / 1.149 R N/kg; mPJE\_lambda… \\
You 2026$^\dagger$ (J Educ Health Sport) & OpenCap $\rightarrow$ whole-body COM acceleration (non-ML) & 18 & n/a: n/a (no training) &  & no & no & r 0.62-0.94; peak landing force bias $>$ 40\% \\
\multicolumn{8}{l}{\textbf{Multi-study and physics-based}} \\
Fluit 2014$^\dagger$ (J Biomech) & Markers $\rightarrow$ AnyBody dynamic contact model (physics) & 9 & n/a: n/a (no training) &  & no & no & no significant differences predicted vs measured except transverse GRM \\
Karatsidis 2017$^\dagger$ (Sensors) & Xsens inertial mocap $\rightarrow$ Newton-Euler + smooth-transition as… & 11 & n/a: n/a (fSTA curves fitted in-sample) &  & yes & no & vertical rho 0.992, rRMSE 5.3\%; lateral rho 0.862, 13.1\%; frontal GRM… \\
Skals 2017$^\dagger$ (Multibody Syst Dyn) & 35 markers $\rightarrow$ AnyBody, 18 contact elements/foot (physics) & 10 & n/a: n/a (no training) &  & yes & no & vertical r median 0.99; ML r median 0.61; transverse GRM median 0.09;… \\
Werling 2023 (AddBiomechanics tool)$^\dagger$ (PLoS ONE) & Markers + measured force plates $\rightarrow$ sequential optimisation… & 10 + Uhlr… & :  &  & partial & no & marker RMSE 1.5 vs 4.3 cm expert; residuals $<$ 2\% of peak external for… \\
Werling 2024 (AddBiomechanics Dataset) (ECCV) & OpenSim joint kinematics from 15 studies / 12 labs (MLP, Gr… & 273 & not stated: not stated ('standard train/dev/test sp… & not stated & no & no & test GRF L2 error: analytical 2.79, MLP 1.29, GroundLinkNet 1.17 N/kg \\
Tan 2025 (GaitDynamics) (Nat Biomed Eng 2025 (…) & OpenSim kinematics (Rajagopal, 75 inputs) from AddBiomechan… & 270 (178… & study: study-level (7 of 15 studies held out)… & yes & yes & yes & MAE 3.2 / 1.2 / 0.7 \%BW (V/AP/ML); held-out studies 1.1-3.9\% vertical… \\
Di Pietro 2025 (OpenGRF)$^\dagger$ (bioRxiv (preprint)) & OpenSim kinematics $\rightarrow$ calibrated contact probes + CMC (phys… & 7 & n/a: n/a (no training) &  & yes & no & vertical nRMSE 1.4\%, rho 0.99 (walking); ML 11.2\%, rho 0.48 (descent)… \\
Golani 2026$^\dagger$ (arXiv (preprint)) & Qualisys markers $\rightarrow$ whole-body CoM acceleration + minimum-r… & 4 (from ~… & n/a: n/a (no training) &  & no & no & vGRF R2 0.967 L / 0.970 R \\
\end{longtable}
\end{center}

\twocolumn

\end{document}